\documentclass[journal]{IEEEtran}

\usepackage{cite}
\usepackage{amsmath,amssymb,amsfonts}
\usepackage{graphicx}
\usepackage{algorithm,algorithmic}
\usepackage{textcomp}
\usepackage{xcolor}
\usepackage{siunitx}
\usepackage{booktabs}
\usepackage{mdframed}
\usepackage{fancyvrb,multirow}
\usepackage{soul}
\usepackage{dsfont,mathabx}
\usepackage{array}
\usepackage{makecell}
\usepackage{diagbox}
\usepackage{threeparttable}
\usepackage{bm}
\usepackage{cuted}
\usepackage{subcaption}
\usepackage{float}
\usepackage{tabularx}
\usepackage{stfloats}
\usepackage{afterpage}
\usepackage{placeins}
\usepackage{tikz}
\usepackage[hidelinks]{hyperref}
\hypersetup{
    pdftitle={DIVAL: Diffeomorphic Vector-Field Alignment for Nonholonomic Motion Planning},
    pdfauthor={Yike Qiao, Xiaodong He, An Zhuo, Zhiyong Sun, Weimin Bao, and Zhongkui Li},
    pdfkeywords={Nonholonomic constraint, motion planning, collision avoidance, vector field}
}
\newcolumntype{C}{>{\centering\arraybackslash}X}
\newtheorem{theorem}{Theorem}
\newenvironment{proof}{\par\noindent\textbf{Proof.}\ }{\hfill$\square$\par}
\newtheorem{proposition}{Proposition}

\newtheorem{assumption}{Assumption}
\newtheorem{definition}{Definition}
\newtheorem{remark}{Remark}

\title{\fontsize{18}{20}\selectfont\bfseries
DIVAL:
Diffeomorphic Vector-Field Alignment for \\Nonholonomic Motion Planning
}

\author{Yike Qiao, Xiaodong He, An Zhuo, Zhiyong Sun, Weimin Bao, and Zhongkui Li}

\begin{document}

\maketitle

\begin{abstract}
    Conventional vector-field-based motion planners for nonholonomic robots consist of two steps: first designing a nominal vector field and then controlling the robot to follow it to approach the target.
    However, a nonholonomic robot cannot follow the vector field instantaneously and may collide with obstacles, due to the inevitable mismatch between the robot's initial heading and the vector field's direction.
    This paper proposes a diffeomorphic vector-field alignment (DIVAL) method, which aligns the vector field's direction with the robot's initial heading so the robot can follow the vector field from the start.
    Specifically, the nominal vector field is locally reshaped by applying a diffeomorphic deformation to the workspace, such that the reshaped vector field has the same direction as the robot's heading at the initial position and coincides with the nominal vector field outside the deformed region.
    Since DIVAL eliminates the initial heading mismatch, the nonholonomic robot can maintain heading alignment with the reshaped vector field throughout the motion.
    This heading alignment further ensures that the robot preserves the nominal vector field's properties, including obstacle avoidance, convergence to the target, and smooth control signals.
    Numerical simulations provide vector-field-tracking baselines where the nominal vector fields are reshaped with DIVAL.
    Compared to the original baselines, the proposed method achieves collision-free convergence in all trials and avoids control discontinuities.
    Hardware experiments on a ground vehicle further demonstrate collision-free convergence across different targets and initial conditions under real-world disturbances.
\end{abstract}


\begin{figure}[!t]
    \centering
    \begin{subfigure}[t]{\columnwidth}
        \centering
        \includegraphics{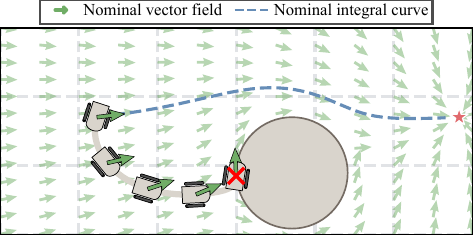}
        \caption{Trajectory under the vector-field-tracking method.}
        \label{fig:overview-baseline}
    \end{subfigure}
    \hfill
    \begin{subfigure}[t]{\columnwidth}
        \centering
        \includegraphics{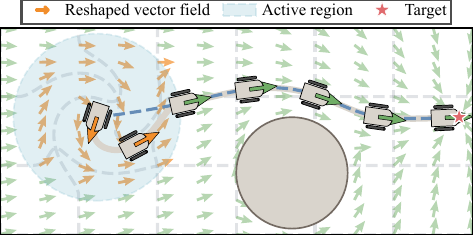}
        \caption{Trajectory under diffeomorphic vector-field alignment (DIVAL).}
        \label{fig:overview-dival}
    \end{subfigure}
    \caption{\textbf{Comparison between the vector-field-tracking method and DIVAL (proposed).}
    (\ref{sub@fig:overview-baseline}): The vector-field-tracking planner gradually rotates the nonholonomic robot towards the vector field's direction. The robot deviates from the integral curve of the vector field due to the initial heading mismatch, which may lead to undesirable behaviors such as collision.
    (\ref{sub@fig:overview-dival}): DIVAL locally reshapes the vector field to align it with the robot's initial heading by deforming the workspace (visualized by the gray dotted curves). Since the heading mismatch is eliminated at the beginning, the robot can follow the reshaped vector field from the start and converge to the target without collision.
    }
    \label{fig: overview}
\end{figure}

\section{Introduction}

Nonholonomic constraints of mobile robots, such as wheeled uncrewed ground vehicles (UGVs) and fixed-wing uncrewed aerial vehicles (UAVs), restrict their admissible motion to the heading direction and pose challenges for motion planning \cite{egeland1996feedback,morro2011path,li2013finite,liu2024ticop,he2024roto,he2024simultaneous}.
Among existing motion planning methodologies, vector-field-based motion planners are suitable for nonholonomic robots because the vector assigned to each position specifies the desired heading direction of the robot \cite{panagou2011control,kapitanyuk2018guiding,feurgard2024extending}.
The prevailing paradigm of vector-field-based planning consists of two stages \cite{masoud2009harmonic,panagou2010dipole,panagou2014motion}.
First, a vector field is designed with desirable properties, including safety, convergence, and smoothness, such that its integral curves avoid obstacles and converge smoothly to the target.
Next, an attitude tracking controller is designed to rotate the robot's heading toward the vector field's direction as it moves forward, so that the robot follows an integral curve once the heading mismatch is stabilized at zero.

One classic way to generate the vector field is computing the gradient of a potential function, such as artificial potential field \cite{Khatib1986Khatib,warren1989global,pathak2005integrated}, harmonic potential field \cite{connolly1993applications,lau2015fluid,wang2025customize}, navigation functions \cite{koditschek1990robot,Rimon1992Exact,tanner2005decentral}, and the data-driven approaches \cite{ni2023ntfields,rousseas2024reactive,liu2025physics}.
Besides the gradient-based methods, some works decompose the environment into polygonal cells and directly design vector fields within each cell \cite{lindemann2009simple,putz2021continuous}.
The aforementioned vector fields are safe, convergent, and smooth along their integral curves.
Thus, when the nonholonomic robot has the same heading as the vector field, it can move along the integral curve and exhibit the prescribed properties.

Since vector fields usually disregard the robot's initial heading, an initial heading mismatch between the robot and the vector field exists in most cases.
To handle this problem, existing works employ the attitude tracking controllers to regulate  the heading mismatch\cite{pothen2017curvature,yao2020path,shivam2021curvature,yao2023guiding}, which can be categorized into two types.
One type is the asymptotic tracking controllers \cite{Panagou2017Distributed,yao2022guiding,he2025novel,qiao2026curvature}, where the robot's heading asymptotically converges to the vector field's direction, so that the robot can move along the vector field's integral curve as the time goes to infinity.
The other type is the finite-time tracking controllers \cite{qiao2024motion,yang2025finite}, where the robot strictly moves along the integral curve once the heading mismatch converges to zero in a finite time.

However, neither controller type eliminates the initial heading mismatch instantaneously, so the heading mismatch persists during the robot's motion.
Since the robot can only move along its heading direction, the heading mismatch inevitably causes the robot's trajectory to deviate from the vector field's integral curve.
This deviation may cause collision with obstacles or discontinuities in control inputs even though the vector field is inherently safe and smooth, and consequently undermine the vector field's desirable properties under the attitude tracking controller.
Representative cases where the robot encounters collisions and fails to converge to the target are illustrated in Figure~\ref{fig: overview}(\ref{sub@fig:overview-baseline}) and numerical simulations in Section~\ref{subsec:simulation}.

Instead of designing attitude tracking controllers for the robot to reduce the heading mismatch, another possible way is to reshape the nominal vector field such that its direction is aligned with the robot's heading at the initial position.
Although such an idea is natural, reshaping the vector field remains challenging, since it involves not only altering the vector field's direction to eliminate the heading mismatch, but also preserving desirable properties of the nominal vector field in terms of safety, convergence, smoothness, among others. Otherwise, the robot may collide with obstacles or cannot reach the destination even though it strictly follows the vector field from the start.

To address the above-mentioned challenges, we propose a diffeomorphic vector-field alignment (DIVAL) framework for motion planning of robots with 3D  nonholonomic kinematics.
DIVAL defines a diffeomorphism according to the initial heading mismatch between the nominal vector field and the nonholonomic robot, and locally deforms the workspace around the robot's initial position.
The workspace deformation induces a corresponding reshaping effect of the nominal vector field, as in Figure~\ref{fig: overview}(\subref{fig:overview-dival}).
At the initial position, the reshaped vector field is aligned with the robot's heading, and hence the initial heading mismatch is eliminated.
As the deformation gradually vanishes away from the initial position, the reshaped vector field smoothly transitions back to the nominal vector field and remains identical to it outside the deformed region.

Furthermore, we design control laws to maintain the initial heading alignment established by DIVAL throughout the robot's motion.
Under the proposed control laws, the robot moves strictly along the integral curve of the reshaped vector field in the ideal case, and within a small bounded neighborhood of that curve under disturbances.
Moreover, given that the deformation of DIVAL is a diffeomorphism, we show that the integral curves of the reshaped vector field are the images of the integral curves of the nominal vector field.
We further prove that under explicit initial conditions, the nonholonomic robot can avoid obstacles, converge to the target, and maintain continuity of control inputs as the nominal vector field.

The theoretical contributions are verified through extensive numerical simulations and hardware experiments.
In simulation, we select several vector-field-tracking baselines for comparison and reshape their nominal vector fields with DIVAL.
Compared with the baselines that rely on aggressive control inputs, DIVAL eliminates trajectory deviation and ensures collision-free convergence for all initial configurations in the given set without abrupt control changes.
The effectiveness of DIVAL in real-world applications is further examined on a UGV, which demonstrates safe convergence covering different target and initial configurations despite disturbances.

This paper is organized as follows. Section~\ref{sec:problem} formulates the objectives of vector-field alignment.
Section~\ref{sec:dival} proposes DIVAL and proves that the reshaped vector field satisfies the alignment objectives.
Section~\ref{sec:planning} derives the control law and shows that the closed-loop system preserves the safety, convergence, and smoothness properties of the nominal vector field.
Section~\ref{sec:validation} presents the simulation results including comparisons with the baselines, and validates the effectiveness of DIVAL in real-world experiments.

\section{Problem Formulation}
\label{sec:problem}

{We consider the kinematic model of a nonholonomic robot moving in 3D Euclidean space.}
The robot's configuration is given by $\bm g = (\bm p, \mathbf{R}) \in SE(3)$, with position $\bm p \in \mathbb{R}^3$ and attitude $\mathbf{R} \in SO(3)$.
{By taking the linear velocity $v \in [v_{\min},v_{\max}]$, $v_{\min} \ge 0$, and the angular velocity ${\bm \Omega} \in \mathbb{R}^3$ as the directly assignable control inputs,} the robot's kinematics are given by
\begin{equation} \label{eq:kinematics}
    \begin{aligned}
        \dot{\bm{p}} &= v \bm{h}, \\
        \dot{\mathbf{R}} &= \mathbf{R} [{\bm \Omega}]_\times,
    \end{aligned}
\end{equation}
where $\bm h \in \mathbb{R}^3$ is the robot's heading direction, given by the first column of $\mathbf{R}$, and $[\cdot]_\times$ is the skew-symmetric matrix operator.
The initial configuration of the robot is denoted by $\bm g_0 = (\bm p_0, \mathbf{R}_0)$, and the initial heading is denoted by $\bm h_0$.

{
The conventional vector-field-tracking approach, as shown in Figure~\ref{fig: overview}(\subref{fig:overview-baseline}), keeps the vector field fixed and uses an attitude tracking controller to gradually rotate the robot toward the vector field's direction.
Due to the heading mismatch during motion, the robot inevitably deviates from the nominal vector field's integral curve, probably leading to collision.
A trivial solution to avoid this deviation is to first rotate the robot in place at $\bm p_0$ until its heading aligns with the nominal vector field and then start moving.
However, many nonholonomic robots, such as fixed-wing UAVs and car-like vehicles, cannot rotate in place.
We therefore exclude this trivial case by requiring $\bm \Omega \to \bm 0$ as $v \to 0$.
Under this circumstance, the robot must begin moving from its given initial heading.}

To address the initial heading mismatch without rotation in place, we aim to reshape the nominal vector field $\bm \xi$ as illustrated in Figure~\ref{fig: overview}(\subref{fig:overview-dival}).
In this way, the reshaped vector field $\bm \eta$ aligns with the robot's initial heading, i.e.,
\begin{equation} \label{eq: align}
    \hat{\bm \eta}(\bm p_0) = \bm h_0,
\end{equation}
where $\hat{\bm \eta}$ denotes the normalization of $\bm \eta$.
This alignment eliminates the initial heading mismatch between the robot and the vector field, allowing the robot to follow the reshaped vector field from the start.

In order to accomplish the alignment without changing the nominal vector field globally, the reshaping should be confined to a neighborhood $\mathcal A\subseteq\mathrm{int}(\mathcal D_{\bm \xi})$ of the initial position $\bm p_0$, where $\mathcal D_{\bm \xi}\subseteq\mathbb R^3$ is the domain of definition for the nominal vector field $\bm \xi$.
Therefore, we require that
\begin{equation} \label{eq: local}
    \bm \eta(\bm p) = \bm \xi(\bm p), \quad \forall \bm p \in \mathcal D_{\bm \xi} \setminus \mathcal A.
\end{equation}

This requirement ensures that outside $\mathcal{A}$, the reshaped vector field is identical to the nominal vector field, and thus can preserve the smoothness of the nominal vector field specified in the following assumption.
\begin{assumption}
    \label{ass:nominal smoothness}
    The nominal vector field $\bm \xi$ has continuous derivatives up to order $r$ on the domain of definition $\mathcal{D}_{\bm \xi}$, i.e., $\bm \xi \in C^r(\mathcal{D}_{\bm \xi})$, for a certain integer $r\ge1$.
\end{assumption}

Specifically, we require the reshaped vector field to have the same smoothness as the nominal one, not only outside $\mathcal{A}$, but on the same domain of definition including $\mathcal{A}$, i.e.,
\begin{equation} \label{eq: smooth}
    \bm \eta \in C^r(\mathcal D_{\bm \eta}), \quad
    \mathcal D_{\bm \eta} = \mathcal D_{\bm \xi}.
\end{equation}
{
Under Assumption~\ref{ass:nominal smoothness} and requirement~\eqref{eq: smooth}, the nominal and reshaped vector fields admit unique integral curves starting from any initial position $\bm p_0\in\mathrm{int}(\mathcal D_{\bm\xi})$ \cite[Theorem~2.2]{teschl2012ordinary}.
Denote these curves by $\psi_{\bm\xi}(t;\bm p_0)$ and $\psi_{\bm\eta}(t;\bm p_0)$, and their respective intervals of existence by $\mathcal T_{\bm\xi},\mathcal T_{\bm\eta}\subseteq\mathbb R_{\ge0}$, both starting at $t=0$.
For $\bm\zeta\in\{\bm\xi,\bm\eta\}$ and $t\in\mathcal T_{\bm\zeta}$, the corresponding integral curve satisfies
\[
\begin{aligned}
    \frac{\mathrm{d}}{\mathrm{d}t}\psi_{\bm\zeta}(t;\bm p_0)
    &=\bm\zeta\!\left(\psi_{\bm\zeta}(t;\bm p_0)\right),\\
    \psi_{\bm\zeta}(0;\bm p_0)&=\bm p_0.
\end{aligned}
\]
}

Note that the nominal and reshaped vector fields are generally different inside $\mathcal A$, and hence their integral curves also become distinct.
To prevent the reshaped integral curve from being trapped in $\mathcal A$ and to ensure that the robot can return to the nominal vector field, we require that it leaves $\mathcal A$ whenever the nominal integral curve does.
This requirement is formalized as
\begin{equation} \label{eq: leave}
    \psi_{\bm \eta}(t;\bm p_0) \notin \mathcal A,
    \quad \forall t \in T_{\neg \mathcal A},
\end{equation}
where $T_{\neg\mathcal A}\triangleq\left\{t\in\mathcal T_{\bm\xi}\mid\psi_{\bm\xi}(t;\bm p_0)\notin\mathcal A\right\}$ is the set of times at which the nominal integral curve lies outside $\mathcal A$.

To summarize, the objective of this paper is to present a vector field alignment method in which the reshaped vector field satisfies the properties~\eqref{eq: align}--\eqref{eq: leave}, such that the robot can strictly move along its integral curve without deviation and smoothly transition back to the nominal vector field.

\section{Diffeomorphic Vector-Field Alignment (DIVAL)}
\label{sec:dival}

This section presents DIVAL and proves that the reshaped vector field satisfies the requirements stated in Section~\ref{sec:problem}.

{
    To align the vector field with the robot's initial heading, we reshape the nominal vector field by deforming the workspace around the robot's initial position. Let $\Phi$ denote this deformation, mapping a position $\bm q$ before deformation to $\bm p=\Phi(\bm q)$ after deformation, where $\bm q$ and $\bm p$ are referred to as the \textit{nominal} and \textit{deformed} positions, respectively. The deformation changes both the positions and the directions and magnitudes of the attached vectors, transforming the nominal vector $\bm\xi(\bm q)$ at $\bm q$ into the reshaped vector $\bm\eta(\bm p)$ at $\bm p=\Phi(\bm q)$.
}

{
    To change the vector direction at the robot's initial position while keeping that position unchanged, we choose $\bm p_0$ as the fixed center of the deformation such that $\bm p_0 = \Phi(\bm q_0)=\bm q_0$.
}
Let $\hat{\bm{\xi}}_0 \triangleq \hat{\bm{\xi}}(\bm p_0)$ denote the nominal vector field's direction at the robot's initial position $\bm p_0$, where $\hat{\bm{\xi}}$ is the normalization of $\bm{\xi}$.
The rotation matrix that rotates $\hat{\bm{\xi}}_0$ to the robot's initial heading $\bm{h}_0$ is given by Rodrigues' rotation formula \cite[Eq. 3.51]{lynch2017modern} as
 \begin{equation} \label{eq: R_*}
\mathbf{R}_* = \mathbf{I} + [\hat{\bm{\xi}}_0 \times \bm{h}_0]_\times + \frac{1}{1+\langle \hat{\bm{\xi}}_0, \bm{h}_0 \rangle}[\hat{\bm{\xi}}_0 \times \bm{h}_0]_\times^2,
\end{equation}
which is well-defined when the nominal vector field is nonzero and not opposite to the robot's initial heading, i.e., $\bm{\xi}(\bm{p}_0) \neq \bm{0}$ and $\langle \hat{\bm{\xi}}_0, \bm{h}_0 \rangle > -1$.
Using the logarithmic map on $SO(3)$, the rotation matrix $\mathbf{R}_*$ can be represented in vector form as
\begin{equation} \label{eq: Delta_*}
    [\bm{\Delta}_*]_\times \triangleq \log_{SO(3)}(\mathbf{R}_*),
\end{equation}
where $\bm{\Delta}_* \in \mathbb{R}^3$.
When $\hat{\bm{\xi}}_0\neq\bm h_0$ and hence $\mathbf R_* \neq \mathbf I$, the direction of $\bm{\Delta}_*$ represents the rotation axis $\hat{\bm{\xi}}_0\times\bm h_0$, and its magnitude $\|\bm \Delta_*\|$ equals the initial heading mismatch angle between $\hat{\bm{\xi}}_0$ and $\bm h_0$.
In particular, when $\hat{\bm{\xi}}_0$ and $\bm h_0$ align, the rotation matrix $\mathbf{R}_*$ is the identity and $\bm{\Delta}_*=\bm 0$.

Recall that the deformation is confined to a neighborhood of the initial position $\bm p_0$. We define this neighborhood as the following active region:
\begin{equation} \label{eq: active region}
    \mathcal{A} \triangleq \{\bm{q}\in\mathcal{D}_{\bm{\xi}} \mid \|\bm{q}-\bm{p}_0\| < r_\mathcal{A}\},
\end{equation}
where $r_\mathcal{A} \le \mathrm{dist}(\bm{p}_0,\partial \mathcal{D}_{\bm{\xi}})$ and $\mathcal{A}$ is a subset of the nominal vector field's domain of definition $\mathcal{D}_{\bm{\xi}}$.
The condition $\mathcal A\subset\mathcal D_{\bm{\xi}}$ ensures that the deformation is applied only where the nominal vector field is defined, so that the workspace boundary determined by obstacles or other environmental constraints remains unchanged.

We then introduce a transition function $\lambda : \mathcal{D}_{\bm{\xi}} \to [0,1]$ to smoothly decay the rotation vector $\bm\Delta_*$ at the initial position $\bm p_0$ to the zero vector outside the active region $\mathcal{A}$.
{
The required properties of the transition function are specified in the following definition.
\begin{definition}
\label{def:admissible transition functions}
For an integer $r\geq 0$ and the active region $\mathcal A$ defined in \eqref{eq: active region} with $r_{\mathcal A}>0$, a function $\lambda:\mathcal D_{\bm\xi}\to[0,1]$ is called an \emph{admissible transition function}, if it is radially symmetric with respect to $\bm p_0$ within $\mathcal A$ and satisfies
\begin{gather*}
    \lambda(\bm p_0)=1,\\
    \lambda(\bm q)=0,
    \quad \forall\,\bm q\in\mathcal D_{\bm\xi}\setminus\mathcal A,\\
    \lambda\in C^{r+1}(\mathcal D_{\bm\xi}).
\end{gather*}
The family of all such functions is denoted by $\Lambda^{r+1}(\bm p_0,r_{\mathcal A})$.
\end{definition}

An example of such a function is a radially symmetric $C^\infty$ bump function built on the squared distance to $\bm p_0$ that vanishes outside $\mathcal A$ and is normalized to attain its maximum value of one at $\bm p_0$.
Consequently, the above family is nonempty for every integer $r\geq 0$.
}
Using the transition function, the rotation vector at each nominal position $\bm q$ is given by
\begin{equation} \label{eq: Delta}
    \bm{\Delta}(\bm{q}) = \lambda(\bm{q}) \bm{\Delta}_*,
\end{equation}
which is equal to $\bm\Delta_*$ at the initial position $\bm p_0$ and smoothly reduces to zero outside the active region $\mathcal{A}$.

Taking the initial position $\bm p_0$ as the center of deformation and applying the rotation $\bm \Delta(\bm q)$ to each point at the nominal position $\bm q$, we move the nominal position $\bm q$ to the deformed position $\bm p$ with the following deformation map $\Phi \colon \mathcal{D}_{\bm{\xi}} \to \mathcal{D}_{\bm{\eta}}$, defined by
\begin{equation} \label{eq: diffeomorphism 3D}
    \bm p = \Phi(\bm{q})= \exp([\bm{\Delta}(\bm{q})]_\times)\bar{\bm{q}} + \bm{p}_0,
\end{equation}
where $\exp([\bm\Delta(\bm q)]_\times)$ maps the rotation vector to a rotation matrix in $SO(3)$, and $\bar{\bm{q}}=\bm{q}-\bm{p}_0$ denotes the displacement with respect to the initial position.
The image set of the deformation map is denoted by
$\mathcal{D}_{\bm{\eta}} \triangleq \Phi(\mathcal{D}_{\bm{\xi}})
= \left\{ \Phi(\bm q) \mid \bm q \in \mathcal{D}_{\bm{\xi}} \right\}$.

Intuitively, the deformation map rotates the point at the nominal position $\bm q$ with respect to the center $\bm p_0$ according to the axis-angle rotation specified by $\bm\Delta(\bm q)$ such that the deformed position $\bm p$ is obtained.
As the nominal position $\bm q$ leaves the active region $\mathcal{A}$, the rotation vector $\bm \Delta(\bm q)$ vanishes, and hence both the rotation matrix and the deformation map reduce to the identity, i.e., $\bm p = \Phi(\bm{q}) = \bm{q}$, $\forall \bm q \in \mathcal{D}_{\bm{\xi}} \setminus \mathcal{A}$.

Via the deformation constructed above, the integral curves of the nominal vector field are deformed accordingly.
Let $\bm q(t) \triangleq \psi_{\bm{\xi}}(t;\bm q(0))$ be a nominal integral curve starting from $\bm q(0) \in \mathcal{D}_{\bm{\xi}}$ so that $\dot{\bm q}=\bm{\xi}(\bm q)$.
The deformation map \eqref{eq: diffeomorphism 3D} transfers each position on the nominal integral curve $\bm q(t)$ to the deformed integral curve $\bm p(t)=\Phi(\bm q(t))$.
Since a vector field gives the velocity along its integral curves, differentiating $\bm p(t)$ defines the reshaped vector field $\bm\eta$:
\begin{equation} \label{eq: aligned vf}
    \begin{aligned}
        \bm{\eta}(\bm p)
        &= \dot{\bm p}
        = \frac{\partial \Phi}{\partial \bm{q}} \dot{\bm{q}}
        = \mathbf{J}_\Phi(\bm{q})\,\bm{\xi}(\bm{q})\\
        &=\mathbf{J}_\Phi\!\left(\Phi^{-1}(\bm p)\right)\bm{\xi}\!\left(\Phi^{-1}(\bm p)\right),\quad \bm p \in \mathcal{D}_{\bm{\eta}},
    \end{aligned}
\end{equation}
where $\Phi^{-1}\colon\mathcal D_{\bm{\eta}}\to\mathcal D_{\bm{\xi}}$ is the inverse of the deformation map, and $\mathbf{J}_\Phi$ is the Jacobian of the deformation map.

In Equation \eqref{eq: aligned vf}, the effects of the deformation on the vector field are illustrated.
Firstly, the deformation transfers each nominal position $\bm q$ to a deformed position $\bm p$.
Therefore, given a deformed position $\bm p$, Equation \eqref{eq: aligned vf} uses the inverse map $\bm q=\Phi^{-1}(\bm p)$ to recover the nominal position in order to compute $\bm{\xi}(\bm q)$.
Secondly, the deformation locally stretches and shears the space around the nominal position $\bm q$ and the nominal vector field $\bm \xi$ is transformed by the Jacobian $\mathbf J_\Phi(\bm q)$.
As a result, the reshaped vector field $\bm \eta$ may have a different magnitude and direction from $\bm \xi$ in the active region $\mathcal{A}$.

To evaluate the reshaped vector field in \eqref{eq: aligned vf}, we need to compute the inverse map $\Phi^{-1}$. Note that the forward map $\Phi$ in \eqref{eq: diffeomorphism 3D} preserves the distance to the deformation center $\bm p_0$, i.e., $\|\bm q - \bm p_0\| = \|\bm p - \bm p_0\|$.
For any $\lambda\in\Lambda^{r+1}(\bm p_0,r_{\mathcal A})$, radial symmetry in Definition~\ref{def:admissible transition functions} gives $\lambda(\bm q)=\lambda(\bm p)$, and it follows from \eqref{eq: Delta} that $\bm\Delta(\bm q)= \bm\Delta(\bm p)$.
Since the forward map $\Phi$ rotates the nominal displacement vector $\bm q-\bm p_0$ by $\exp([\bm\Delta(\bm q)]_\times)$, the inverse map can be obtained by applying the inverse rotation $\exp([\bm\Delta(\bm p)]_\times)^\intercal$ to the deformed displacement vector $\bm p - \bm p_0$.
In this way, the inverse map is explicitly given by
\begin{equation} \label{eq: inv}
    \Phi^{-1}(\bm{p}) = \exp([\bm{\Delta}(\bm{p})]_\times)^\intercal (\bm p - \bm p_0) + \bm{p}_0.
\end{equation}
Once $\bm q = \Phi^{-1}(\bm p)$ is obtained, the Jacobian $\mathbf{J}_\Phi(\bm q)$ can be further computed by differentiating \eqref{eq: diffeomorphism 3D} with respect to $\bm q$ as follows:
\begin{equation} \label{eq: jac}
    \mathbf{J}_\Phi(\bm{q}) = \exp([\bm{\Delta}(\bm{q})]_\times)
    \left(\mathbf{I} + [\bm{\Delta}_*]_\times \bar{\bm{q}} \nabla \lambda(\bm{q})^\intercal\right).
\end{equation}
The detailed derivation of \eqref{eq: inv} and \eqref{eq: jac} is provided in Appendix~\ref{appendix: eta derivation}.

In the above construction of DIVAL, we have obtained the reshaped vector field $\bm{\eta}$ from the nominal vector field $\bm{\xi}$ by the deformation map $\Phi$.
{To ensure that the reshaped vector field is well-defined, we impose the following conditions on the robot's initial configuration.
In \eqref{eq: Delta_*}, the rotation vector $\bm{\Delta}_*$ requires a non-singular nominal vector at the initial position and excludes the antiparallel case\footnote{\label{fn:antiparallel case} The antiparallel case is excluded from the theoretical development because the rotation axis is not unique. In practice, this case can be handled by choosing any rotation vector $\bm\Delta_*$ satisfying $\langle\bm\Delta_*,\hat{\bm\xi}_0\rangle=0$ and $\|\bm\Delta_*\|=\pi$.},
i.e., $\bm{\xi}(\bm p_0)\neq \bm 0$ and $\langle \hat{\bm{\xi}}(\bm p_0),\bm h_0\rangle>-1$.
In \eqref{eq: active region}, the active region $\mathcal A$ requires a positive radius satisfying
$0< r_\mathcal A\leq \mathrm{dist}(\bm p_0,\partial\mathcal D_{\bm{\xi}})$.
Such a radius exists only if $\bm p_0$ lies strictly inside the domain of definition rather than on its boundary, i.e., $\bm p_0\in\mathrm{int}(\mathcal D_{\bm{\xi}})$, which is equivalent to $\mathrm{dist}(\bm p_0,\partial\mathcal D_{\bm{\xi}})>0$.
These conditions define the admissible initial configurations for the DIVAL construction.}
\begin{definition}
\label{def:admissible initial set}
    An initial configuration $\bm g=(\bm p,\mathbf R)\in SE(3)$ is called \emph{admissible} for the DIVAL construction if it belongs to the set
    \begin{equation*}
    \mathcal G_a\triangleq
    \left\{
    \bm g=(\bm p,\mathbf R)\in SE(3)
    \,\middle|\,
    \begin{gathered}
    \bm p \in \mathrm{int}(\mathcal D_{\bm{\xi}}),\\
    \bm{\xi}(\bm p)\neq \bm 0,\\
    \langle \bm h,\hat{\bm{\xi}}(\bm p)\rangle>-1
    \end{gathered}
    \right\},
    \end{equation*}
    where $\bm h$ is the first column of $\mathbf R$.
\end{definition}

The following theorem shows that the requirements in Section~\ref{sec:problem} are satisfied for any initial configuration in $\mathcal G_a$.
\begin{theorem}
    \label{thm:aligned vf}
    Let $\bm \xi$ be a nominal vector field satisfying Assumption~\ref{ass:nominal smoothness}. Given an initial configuration $\bm g_0=(\bm p_0,\mathbf R_0)\in\mathcal G_a$, for any active-region radius $0 < r_{\mathcal A} \le \operatorname{dist}(\bm p_0,\partial\mathcal D_{\bm \xi})$ and {any transition function $\lambda \in \Lambda^{r+1}(\bm p_0,r_{\mathcal A})$ in Definition~\ref{def:admissible transition functions},} the resulting deformation map $\Phi$ defined by \eqref{eq: diffeomorphism 3D} is a $C^{r+1}$ diffeomorphism, and the reshaped vector field $\bm \eta$ defined by \eqref{eq: aligned vf} satisfies the properties \eqref{eq: align}--\eqref{eq: leave}.
\end{theorem}
\begin{proof}
    We first verify that $\Phi$ in \eqref{eq: diffeomorphism 3D} is a diffeomorphism and show that the deformed domain $\mathcal D_{\bm{\eta}}$ is identical to the nominal domain $\mathcal D_{\bm{\xi}}$.
    Since $\bm g_0\in\mathcal G_a$, the rotation vector $\bm\Delta_*$ is well-defined, and the radius constraint $0 < r_{\mathcal A} \le \operatorname{dist}(\bm p_0,\partial\mathcal D_{\bm \xi})$ ensures that $\mathcal A\subseteq\mathcal D_{\bm\xi}$.
    Given the inverse map of \eqref{eq: diffeomorphism 3D} in \eqref{eq: inv}, a direct substitution yields that $\Phi^{-1}(\Phi(\bm q))=\bm q$ and $\Phi(\Phi^{-1}(\bm p))=\bm p$, and hence $\Phi$ is a bijection.
    It follows from $\lambda\in C^{r+1}(\mathcal D_{\bm{\xi}})$ that both $\Phi$ and $\Phi^{-1}$ are of class $C^{r+1}$.
    Therefore, $\Phi$ is a $C^{r+1}$ diffeomorphism.
    In addition, since the diffeomorphism $\bm p = \Phi(\bm q)$ only rotates the nominal position $\bm q$ around the initial position $\bm p_0$, it preserves the distance to $\bm p_0$, i.e.,
    \[
    \|\bm p-\bm p_0\|=\|\bm q-\bm p_0\|.
    \]
    Since $\mathcal A$ is a ball centered at $\bm p_0$, distance preservation means that a nominal position $\bm q$ inside $\mathcal A$ is mapped to a deformed position $\bm p$ inside $\mathcal A$, so $\Phi(\mathcal A)\subseteq\mathcal A$.
    The converse inclusion $\mathcal A\subseteq\Phi(\mathcal A)$ follows from the bijectivity of $\Phi$, and thereby, $\Phi(\mathcal A)=\mathcal A$.
    Outside $\mathcal A$, the transition function makes the rotation vector vanish, so $\bm p = \Phi(\bm q)=\bm q$ for all $\bm q\in\mathcal D_{\bm{\xi}}\setminus\mathcal A$, which means that $\Phi(\mathcal D_{\bm{\xi}}\setminus\mathcal A)=\mathcal D_{\bm{\xi}}\setminus\mathcal A$.
    Therefore, the deformed domain is unchanged, i.e.,
    \begin{equation} \label{eq: equiv domain}
    \begin{aligned}
        \mathcal D_{\bm{\eta}} &= \Phi(\mathcal D_{\bm{\xi}}) = \Phi(\mathcal A \cup (\mathcal D_{\bm{\xi}} \setminus \mathcal A)) \\ &= \Phi(\mathcal A) \cup \Phi(\mathcal D_{\bm{\xi}}\setminus\mathcal A) = \mathcal D_{\bm{\xi}}.
    \end{aligned}
    \end{equation}

    Next, we demonstrate that the reshaped vector field satisfies the alignment requirement \eqref{eq: align}.
    By substituting the deformed position $\bm p = \bm p_0$ into \eqref{eq: inv}, and then plugging the result into \eqref{eq: jac}, we have
    \begin{equation} \label{eq: value at p0}
    \begin{gathered}
        \Phi^{-1}(\bm p_0)=\bm p_0,\\
        \mathbf J_\Phi(\Phi^{-1}(\bm p_0))=\exp([\bm\Delta_*]_\times)=\mathbf R_*.
    \end{gathered}
    \end{equation}
    Equation~\eqref{eq: value at p0} implies that the diffeomorphism $\Phi$ maps the initial position $\bm p_0$ to itself, and its Jacobian $\mathbf{J}_\Phi$ at $\bm p_0$ is the rotation matrix $\mathbf{R}_*$ given in \eqref{eq: R_*}.
    Evaluating the reshaped vector field \eqref{eq: aligned vf} at $\bm p_0$ yields
    \[
    \begin{aligned}
    \hat{\bm{\eta}}(\bm p_0)
    &= \frac{\mathbf{J}_\Phi(\Phi^{-1}(\bm p_0)) \bm{\xi}(\Phi^{-1}(\bm p_0))}{\|\mathbf{J}_\Phi(\Phi^{-1}(\bm p_0)) \bm{\xi}(\Phi^{-1}(\bm p_0))\|}\\
    &= \mathbf R_*\frac{\bm{\xi}(\bm p_0)}{\|\bm{\xi}(\bm p_0)\|}
    = \mathbf R_*\hat{\bm{\xi}}_0
    = \bm h_0,
    \end{aligned}
    \]
    where the last equality comes from the definition of $\mathbf R_*$ in \eqref{eq: R_*}.
    Therefore, the reshaped vector field aligns with the initial heading at $\bm p_0$, which proves \eqref{eq: align}.

    To verify the requirement of local deformation in \eqref{eq: local}, we consider any position outside the active region $\mathcal{A}$.
    Since $\mathcal D_{\bm{\eta}}=\mathcal D_{\bm{\xi}}$ is established in \eqref{eq: equiv domain}, and both $\lambda$ and $\nabla\lambda$ vanish on $\mathcal D_{\bm{\xi}}\setminus\mathcal A$, it follows from \eqref{eq: inv} and \eqref{eq: jac} that
    \begin{equation} \label{eq: value outside}
    \begin{gathered}
        \Phi^{-1}(\bm p)=\bm p, \quad  \forall \bm p \in \mathcal{D}_{\bm{\eta}} \setminus \mathcal{A} = \mathcal{D}_{\bm{\xi}} \setminus \mathcal{A},\\
        \mathbf J_\Phi(\Phi^{-1}(\bm p))=\mathbf I, \quad \forall \bm p \in \mathcal{D}_{\bm{\eta}} \setminus \mathcal{A} = \mathcal{D}_{\bm{\xi}} \setminus \mathcal{A}.
    \end{gathered}
    \end{equation}
    Substituting \eqref{eq: value outside} into \eqref{eq: aligned vf} gives
    \[
    \bm{\eta}(\bm p)=\bm{\xi}(\bm p),
    \qquad \forall \bm p\in\mathcal D_{\bm{\xi}}\setminus\mathcal A,
    \]
    which proves \eqref{eq: local}.

    Next, the smoothness condition \eqref{eq: smooth} follows from the smoothness of the diffeomorphism and the nominal vector field.
    Since $\Phi$ has been shown to be of class $C^{r+1}$, its Jacobian $\mathbf J_\Phi$ is of class $C^r$.
    Since $\bm{\xi}\in C^r(\mathcal D_{\bm{\xi}})$ by Assumption~\ref{ass:nominal smoothness}, the expression for $\bm{\eta}$ in \eqref{eq: aligned vf} is a composition and product of $C^r$ functions.
    Thus, $\bm{\eta}\in C^r(\mathcal D_{\bm{\eta}})$.
    Together with the domain equality established in \eqref{eq: equiv domain}, this proves \eqref{eq: smooth}.

    Finally, we show that the deformed integral curve, i.e., the integral curve of the reshaped vector field, leaves the active region as required in \eqref{eq: leave}.
    Geometrically, the deformed integral curve $\psi_{\bm{\eta}}$ is the image of the nominal integral curve $\psi_{\bm{\xi}}$ under $\Phi$.
    This relation between the integral curves under a diffeomorphism is stated in \cite[Proposition 9.13]{lee2012introduction} and can be formulated as
    \[
    \psi_{\bm{\eta}}\bigl(t;\Phi(\bm p_0)\bigr)=\Phi\bigl(\psi_{\bm{\xi}}(t;\bm p_0)\bigr).
    \]
    Since \eqref{eq: value at p0} gives $\Phi(\bm p_0)=\bm p_0$, the above relation becomes
    \begin{equation} \label{eq: image}
        \psi_{\bm{\eta}}(t;\bm p_0)=\Phi\bigl(\psi_{\bm{\xi}}(t;\bm p_0)\bigr).
    \end{equation}
    Now let $t'\in T_{\neg\mathcal A}$, where $T_{\neg\mathcal A}$ is defined in \eqref{eq: leave} as the set of time instants when the nominal integral curve lies outside the active region, i.e., $\psi_{\bm{\xi}}(t';\bm p_0)\notin\mathcal A$.
    Since $\Phi(\bm q)=\bm q$ for every $\bm q\in\mathcal D_{\bm{\xi}}\setminus\mathcal A$, we can obtain that
    \begin{equation*}
    \psi_{\bm{\eta}}(t';\bm p_0)=\Phi\bigl(\psi_{\bm{\xi}}(t';\bm p_0)\bigr)=\psi_{\bm{\xi}}(t';\bm p_0),\ \forall t'\in T_{\neg\mathcal A}.
    \end{equation*}
    Therefore, the deformed integral curve also lies outside $\mathcal A$ for every $t'\in T_{\neg\mathcal A}$, which proves \eqref{eq: leave}.
\end{proof}

According to Theorem~\ref{thm:aligned vf}, the reshaped vector field $\bm{\eta}$ satisfies the objectives of vector-field alignment in \eqref{eq: align}--\eqref{eq: leave} for any initial configuration in $\mathcal{G}_a$. Therefore, we refer to $\bm{\eta}$ as the \emph{aligned vector field} and refer to its integral curves $\psi_{\bm{\eta}}$ as \emph{aligned integral curves}.

\begin{figure}[ht]
    \centering
    \includegraphics[width=0.95\columnwidth]{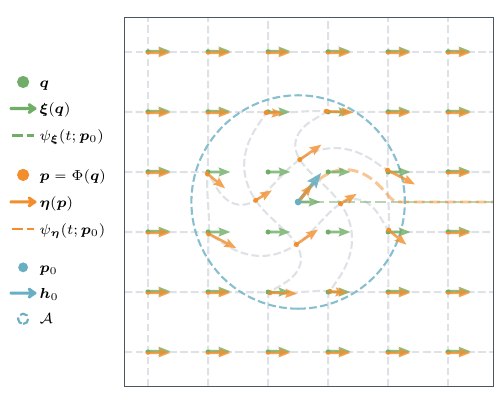}
    \caption{Illustration of 2D DIVAL, with the nominal vector field $\bm \xi=[1,0]^\intercal$.
    The deformation map $\Phi$ maps the uniformly sampled nominal positions $\bm q$ to the deformed positions $\bm p$, inducing the aligned vector field $\bm \eta$ that satisfies $\bm \eta(\bm p_0) = \bm h_0$.
    As $\Phi$ decays to the identity map away from $\bm p_0$, the vector fields $\bm \xi$ and $\bm \eta$ coincide outside the active region $\mathcal A$, as do their integral curves $\psi_{\bm \xi}(t;\bm p_0)$ and $\psi_{\bm \eta}(t;\bm p_0)$.}
    \label{fig:2d reshaping}
\end{figure}

For DIVAL in 2D, planar motion is considered as a special case of the above 3D results, where the third components of the position and vector fields are always zero.
Moreover, the rotation vector \eqref{eq: Delta_*} is reduced to $\bm{\Delta}_* = [0, 0, \delta_*]^\intercal$ in the planar case, where $\delta_*$ is the signed planar rotation angle from the nominal direction $\hat{\bm{\xi}}_0$ to the initial heading $\bm{h}_0$, representing the initial heading mismatch.
Based on the above simplification, the aligned vector field still follows \eqref{eq: aligned vf}, while the diffeomorphism-related quantities are given by
\begin{equation} \label{eq: diffeomorphism 2D}
    \begin{gathered}
        \Phi(\bm{q}) = \exp([{\delta}(\bm{q})]_\times) (\bm{q} - \bm{p}_0) + \bm{p}_0,\\
        \Phi^{-1}(\bm{p}) = \exp([{\delta}(\bm{p})]_\times)^\intercal (\bm{p} - \bm{p}_0) + \bm{p}_0,\\
        \mathbf{J}_\Phi (\bm{q}) = \exp([{\delta}(\bm{q})]_\times) (\mathbf{I} + [{\delta}_*]_\times \bar{\bm{q}} \nabla \lambda^\intercal),\\
    \end{gathered}
\end{equation}
where $\bm q, \bm p \in \mathbb{R}^2$, $\delta(\bm q) = \lambda(\bm q) \delta_* \in \mathbb{R}$, and $\exp([{\delta}(\bm{q})]_\times)$ gives the rotation matrix with rotation angle $\delta(\bm q)$ in 2D.
In Figure~\ref{fig:2d reshaping}, an example of DIVAL in 2D is visualized.

\begin{remark}
    \label{remark:relation to DS mod}
    The methods in \cite{khansari2012dynamical,huber2019avoidance,huber2022avoiding,huber2024rotation} directly alter the vector field at each point to solve obstacle avoidance in a setting where the velocity is directly assignable and every direction is admissible, so the robot's orientation never enters the design.
    By contrast, DIVAL alters the vector field through a workspace deformation to resolve the heading mismatch that every nonholonomic robot faces when the planned vector field is not aligned with its initial attitude. Since a nonholonomic robot admits only the velocities along its heading and can turn that heading only while moving forward, such a mismatch is common. DIVAL resolves it before the motion starts, so that any nominal vector field designed for a fully actuated model becomes executable as it stands.
\end{remark}

\section{Nonholonomic Motion Planning Based on DIVAL}
\label{sec:planning}

With the aligned vector field constructed in Section~\ref{sec:dival}, the remaining question is how the nonholonomic robot can follow this aligned vector field while preserving the prescribed properties of the nominal vector field.
In this section, we first propose control laws under which the robot's heading remains aligned with the vector-field direction throughout the motion and establish the relation between the closed-loop trajectory and the corresponding aligned integral curve.
Based on this relation, we further provide the initial conditions under which the closed-loop system preserves the safety, convergence, and smoothness properties of the nominal vector field.

\subsection{Control Law and Closed-Loop Trajectory}
\label{subsec:control_motion}

Since the aligned vector field has the same direction as the initial heading of the nonholonomic robot at the initial position, the control objective is to preserve this heading alignment during the robot's motion.
Specifically, we require that
\begin{equation}\label{eq: control objective}
    \bm h(t)=\hat{\bm\eta}(\bm p(t))
    =
    \frac{\bm\eta(\bm p(t))}{\|\bm\eta(\bm p(t))\|},
    \quad \forall t\ge0,
\end{equation}
where $\bm h(t)$ is the direction of the robot's heading and $\hat{\bm \eta}$ denotes the direction of the aligned vector field.
Since DIVAL ensures $\bm h_0=\hat{\bm\eta}(\bm p_0)$, we then design the control inputs such that the robot's heading turns with the vector field direction, i.e., $\dot{\bm h}=\dot{\hat{\bm \eta}}$, thereby satisfying \eqref{eq: control objective}.

At any non-singular position of the aligned vector field, i.e., $\bm \eta(\bm p)\neq \bm 0$, we choose the robot's linear velocity as
\begin{equation} \label{eq: v}
    v(\bm g)=k(\bm g)\|\bm \eta(\bm p)\|.
\end{equation}
{In the above equation, $k$ is a speed scaling factor that adjusts the robot's speed relative to the magnitude of the aligned vector field.
It is required to be a class $C^{r-1}$ function of the configuration $\bm g$, satisfying $0<k_{\min}\le k(\bm g)\le k_{\max}<\infty$.
}

The translational kinematics in \eqref{eq:kinematics} then become
\begin{equation} \label{eq: dot p}
    \dot{\bm p}
    =
    v(\bm g)\bm h
    =
    k(\bm g)\|\bm \eta(\bm p)\|\bm h.
\end{equation}
Under this linear velocity, the direction of the aligned vector field evolves as
\begin{equation} \label{eq: dot direction}
    \dot{\hat{\bm \eta}}
    =
    \frac{\partial \hat{\bm \eta}}{\partial \bm p}\dot{\bm p}
    =
    k(\bm g)
    \frac{\partial \hat{\bm \eta}}{\partial \bm p}
    \|\bm \eta(\bm p)\|\bm h.
\end{equation}
In \eqref{eq: dot direction}, $\partial \hat{\bm \eta} / \partial \bm p$ is obtained by differentiating $\hat{\bm \eta}=\bm \eta/\|\bm \eta\|$ with respect to $\bm p$, and the result is given by
\begin{equation} \label{eq: direction jacobian}
    \frac{\partial \hat{\bm \eta}}{\partial \bm p}(\bm p)
    =
    \frac{
        \big(
            \mathbf I
            -
            \hat{\bm \eta}(\bm p)\hat{\bm \eta}^{\intercal}(\bm p)
        \big)
        \mathbf J_{\bm \eta}(\bm p)
    }{
        \|\bm \eta(\bm p)\|
    },
\end{equation}
where $\mathbf J_{\bm \eta}$ is the Jacobian of the aligned vector field $\bm \eta$.
Substituting \eqref{eq: dot p} and \eqref{eq: direction jacobian} into \eqref{eq: dot direction} gives the time derivative of the aligned vector-field direction as follows:
\begin{equation} \label{eq: eta direction rate}
    \dot{\hat{\bm \eta}}
    =
    k(\bm g)
    \big(
        \mathbf I
        -
        \hat{\bm \eta}(\bm p)\hat{\bm \eta}^{\intercal}(\bm p)
    \big)
    \mathbf J_{\bm \eta}(\bm p)\bm h.
\end{equation}

Having obtained $\dot{\hat{\bm \eta}}$, we next derive the changing rate of the robot's heading $\dot{\bm h}$ and design the angular velocity so that the control objective \eqref{eq: control objective} holds. From \eqref{eq:kinematics}, the kinematics of the robot's heading are given by
\begin{equation} \label{eq: dot h}
\begin{aligned}
    \dot{\bm h}
    &=
    \dot{\mathbf R}\bm e_1
    =
    \mathbf R[{\bm \Omega}]_\times\bm e_1 \\
    &=
    \mathbf R({\bm \Omega}\times\bm e_1)
    =
    {\bm \Omega}_{\mathrm{world}}\times(\mathbf{R} \bm e_1)\\
    &= {\bm \Omega}_{\mathrm{world}}\times\bm h,
\end{aligned}
\end{equation}
where $\bm e_1 = [1,0,0]^\intercal$ is the first column of the identity matrix, and ${\bm \Omega}_{\mathrm{world}}\triangleq\mathbf R{\bm \Omega}$ denotes the angular velocity expressed in the world frame.

To establish \eqref{eq: control objective} by enforcing $\dot{\bm h}=\dot{\hat{\bm \eta}}$, the angular velocity ${\bm \Omega}_\mathrm{world}$ should solve the following equation:
\begin{equation} \label{eq: angular velocity equation}
\begin{aligned}
    {\bm \Omega}_{\mathrm{world}}\times\hat{\bm \eta}
    &=
    \left.\dot{\hat{\bm \eta}}\right|_{\bm h=\hat{\bm \eta}} \\
    &=
    k(\bm g)
    \big(
        \mathbf I
        -
        \hat{\bm \eta}(\bm p)\hat{\bm \eta}^{\intercal}(\bm p)
    \big)
    \mathbf J_{\bm \eta}(\bm p)\hat{\bm \eta}(\bm p).
\end{aligned}
\end{equation}
This equation is obtained by substituting \eqref{eq: control objective} into \eqref{eq: eta direction rate} and \eqref{eq: dot h}.
Notice that the matrix $\mathbf I-\hat{\bm \eta}\hat{\bm \eta}^{\intercal}$ projects the vector $\mathbf J_{\bm \eta}\hat{\bm \eta}$ onto the plane orthogonal to $\hat{\bm \eta}$, and hence $\left.\dot{\hat{\bm \eta}}\right|_{\bm h=\hat{\bm \eta}}$ is perpendicular to $\hat{\bm \eta}$.
Since the left-hand side of \eqref{eq: angular velocity equation} can generate any vector perpendicular to $\hat{\bm \eta}$, including $\left.\dot{\hat{\bm \eta}}\right|_{\bm h=\hat{\bm \eta}}$, there exists an angular velocity ${\bm \Omega}_\mathrm{world}$ that satisfies \eqref{eq: angular velocity equation}.

In addition, on the left-hand side of \eqref{eq: angular velocity equation}, any component of ${\bm \Omega}_{\mathrm{world}}$ parallel to $\hat{\bm \eta}$ vanishes in ${\bm \Omega}_{\mathrm{world}}\times\hat{\bm \eta}$ and does not affect the heading motion.
To ensure that the angular velocity solution has the minimum magnitude, we choose ${\bm \Omega}_{\mathrm{world}}\perp\hat{\bm \eta}$.
Then, taking the left cross product of both sides of \eqref{eq: angular velocity equation} with $\hat{\bm \eta}$ gives
\begin{equation*}
    {\bm \Omega}_{\mathrm{world}}
    =
    \hat{\bm \eta}\times({\bm \Omega}_{\mathrm{world}}\times\hat{\bm \eta})
    =
    \hat{\bm \eta}\times
    \left.\dot{\hat{\bm \eta}}\right|_{\bm h=\hat{\bm \eta}}.
\end{equation*}
Therefore, the minimum angular velocity is
\begin{equation} \label{eq: Omega}
    {\bm \Omega}(\bm g)
    =
    \mathbf{R}^\intercal {\bm \Omega}_{\mathrm{world}}
    =
    k(\bm g)\mathbf R^\intercal
    \frac{
        \bm \eta(\bm p)
        \times
        \mathbf J_{\bm \eta}(\bm p)\bm \eta(\bm p)
    }{
        \|\bm \eta(\bm p)\|^2
    }.
\end{equation}
The detailed calculation of \eqref{eq: Omega} is presented in Appendix~\ref{appendix:control_inputs}.

The control laws \eqref{eq: v} and \eqref{eq: Omega} fulfill the control objective \eqref{eq: control objective}, and therefore the closed-loop trajectory $\bm p(t ; \bm g_0)$ evolves on the same geometric path as the aligned integral curve $\psi_{\bm \eta}(t;\bm p_0)$. This relation between the closed-loop trajectory and the aligned integral curve is summarized in the following theorem.

\begin{theorem}
    \label{thm: trajectory relation}
    Consider an initial configuration $\bm g_0=(\bm p_0,\mathbf R_0)\in\mathcal G_a$ and an aligned vector field $\bm \eta$ constructed under the conditions of Theorem~\ref{thm:aligned vf}.
    Let $\mathcal T_{\bm\eta}\subseteq\mathbb R_{\ge0}$ be a time interval starting at zero on which the aligned integral curve $\psi_{\bm\eta}(\cdot;\bm p_0)$ is defined.
    The closed-loop trajectory $\bm p(t;\bm g_0)$ under the control laws \eqref{eq: v} and \eqref{eq: Omega} is a time reparameterization of this integral curve, i.e.,
    \begin{equation} \label{eq: trajectory relation}
            \bm p(t;\bm g_0)
            =\psi_{\bm{\eta}}(\tau(t);\bm p_0),
            \quad \forall\,\tau(t)\in\mathcal T_{\bm\eta},
    \end{equation}
    where
    \begin{equation} \label{eq: time reparam}
        \tau(t)=\int_0^t k(\bm g(s;\bm g_0))\,\mathrm ds.
    \end{equation}
\end{theorem}
\begin{proof}
    See Appendix~\ref{appendix:proof_trajectory_relation}.
\end{proof}

\begin{remark}
    In the vector-field-tracking approaches \cite{Panagou2017Distributed,he2025novel,yao2020path,yao2022guiding,pothen2017curvature,shivam2021curvature,yao2023guiding}, the nonholonomic robot deviates from the integral curve whenever an initial heading mismatch exists. Therefore, the relation between the closed-loop trajectory and the integral curve of the vector field remains unclear.
    In contrast, DIVAL removes the initial heading mismatch by locally deforming the vector field, allowing the robot to maintain the same heading direction as the aligned vector field throughout the motion under the control laws \eqref{eq: v} and \eqref{eq: Omega}.
    Consequently, the closed-loop system moves strictly along the integral curve of the aligned vector field.
\end{remark}

In the planar case, the nonholonomic kinematics \eqref{eq:kinematics} reduce to
\begin{equation} \label{eq:kinematics 2d}
\begin{aligned}
    \dot{\bm p}&=v\bm h,\\
    \dot{\theta}&=\omega,
\end{aligned}
\end{equation}
where $\bm h = [\cos\theta,\sin\theta]^\intercal$ is the heading direction, $\theta$ is the heading angle, $v\in \mathbb{R}_{\ge 0}$ is the linear velocity, and $\omega \in \mathbb{R}$ is the angular velocity.
In the following, we present the planar control law by viewing the 2D system as a special case of the aforementioned 3D construction, where the robot motion and the aligned vector field are confined to the $x$-$y$ plane.
The linear velocity keeps the same form as \eqref{eq: v}, i.e.,
\begin{equation} \label{eq: v 2d}
    v(\bm g)
    =
    k(\bm g)\|\bm \eta(\bm p)\|,
\end{equation}
where the scaling factor $k$ depends on the planar configuration $\bm g = (\bm p, \theta)$ and the aligned vector field $\bm \eta$ is constructed via DIVAL in 2D with the deformation map given by \eqref{eq: diffeomorphism 2D}.

For the angular velocity, when the robot's heading is restricted to the $x$-$y$ plane, the rotation matrix $\mathbf{R}$ in \eqref{eq: Omega} has $[0,0,1]^\intercal$ as its third column.
Meanwhile, the 3D vectors $\bm \eta(\bm p)$ and $\mathbf J_{\bm \eta}(\bm p)\bm \eta(\bm p)$ also lie in the $x$-$y$ plane, so only the third component of their cross product can be nonzero.
Consequently, the 3D angular velocity ${\bm \Omega}$ in \eqref{eq: Omega} can have a nonzero component only along the third axis.
This component corresponds to the scalar angular velocity $\omega$ in \eqref{eq:kinematics 2d}, that is, ${\bm \Omega} = [0,0,\omega]^\intercal$.
For the aligned vector field $\bm \eta \in \mathbb{R}^2$, the angular velocity $\omega$ is given by
\begin{equation} \label{eq: omega 2d}
    \omega(\bm g)
    =
    k(\bm g)
    \frac{
        \det(
            \bm \eta(\bm p),
            \mathbf J_{\bm \eta}(\bm p)\bm \eta(\bm p)
        )
    }{
        \|\bm \eta(\bm p)\|^2
    },
\end{equation}
where $\det(\bm a,\bm b)$ denotes the determinant of the $2\times2$ matrix with $\bm a$ and $\bm b$ as its columns.
Under the control inputs \eqref{eq: v 2d} and \eqref{eq: omega 2d}, the planar closed-loop trajectory $\bm{p}(t; \bm g_0) \in \mathbb{R}^2$ and the aligned integral curve $\psi_{\bm \eta}(t; \bm p_0) \in \mathbb{R}^2$ satisfy the same relation as in Theorem~\ref{thm: trajectory relation}.

\subsection{Property Preservation of the Closed-Loop System}
\label{subsec:guarantees}

In the proof of Theorem~\ref{thm:aligned vf}, Equation \eqref{eq: image} reveals that the aligned integral curve is the image of the nominal integral curve under the deformation map, namely, $\psi_{\bm \eta}(t;\bm p_0) = \Phi(\psi_{\bm \xi}(t;\bm p_0))$.
In addition, Theorem~\ref{thm: trajectory relation} indicates that the closed-loop trajectory is a time reparameterization of the aligned integral curve, i.e., $\bm p(t;\bm g_0)=\psi_{\bm{\eta}}(\tau(t);\bm p_0)$.
Combining these results relates the closed-loop trajectory to the nominal integral curve:
\begin{equation*} \label{eq: relation}
    \bm p(t;\bm g_0)
    =\Phi\left(\psi_{\bm{\xi}}(\tau(t);\bm p_0)\right),
    \ \forall t\ge0.
\end{equation*}
This indicates that the closed-loop trajectory evolves along the image of the nominal integral curve, but with a different time parameterization.
Based on this relation, we demonstrate in the following that the safety, convergence, and smoothness properties of the nominal vector field are preserved by the closed-loop system.

\subsubsection{Safety and Convergence}
Consider a workspace $\mathcal W$ containing obstacles $\mathcal O_i$, $i=1,\ldots,N$, and denote their union by $\mathcal O\triangleq\bigcup_{i=1}^{N}\mathcal O_i$.
The free space is defined as $\mathcal F\triangleq\mathcal W\setminus\mathcal O$, with $\mathcal F\subseteq\mathcal D_{\bm\xi}$ so the nominal vector field is defined on the free space.
Let $\mathcal L_{\bm\xi}^{+}$ denote the target positive limit set, representing either a target equilibrium or a stable limit cycle, and let $\mathcal R_{\bm\xi}^{a}\subseteq\mathcal F$ denote its region of attraction.
To ensure that the angular velocity vanishes with the linear velocity as the robot approaches an equilibrium target, we also require the curvature $\kappa_{\bm\xi}$ of each nominal integral curve to remain finite during convergence.
The following assumption formalizes these requirements.

\begin{assumption}
    \label{ass:nominal safety and convergence}
    The nominal vector field satisfies the following conditions.\\
    \emph{(i) Safety:} any nominal integral curve $\psi_{\bm\xi}(t;\bm p_0)$ starting in $\mathcal F$ remains safe by
    staying in $\mathcal F$, i.e.,
    \begin{equation*}\label{eq: nominal safety}
        \psi_{\bm\xi}(t;\bm p_0)\in\mathcal F,
        \quad\forall t\ge0,\quad\forall\bm p_0\in\mathcal F.
    \end{equation*}
    \emph{(ii) Convergence:} for initial positions in $\mathcal R_{\bm\xi}^{a}$, the nominal
    integral curves converge to the target limit set $\mathcal L_{\bm\xi}^{+}$, i.e.,
    \begin{equation*}\label{eq: nominal convergence}
        \lim_{t\to\infty}\mathrm{dist}\left(\psi_{\bm\xi}(t;\bm p_0),\mathcal L_{\bm\xi}^{+}\right)=0,
        \quad\forall\bm p_0\in\mathcal R_{\bm\xi}^{a}.
    \end{equation*}
    \emph{(iii) Curvature:} along each nominal integral curve with
    $\bm p_0\in\mathcal R_{\bm\xi}^{a}$ and $\bm\xi(\bm p_0)\neq\bm0$, the curvature remains
    finite, i.e.,
    \begin{equation}\label{eq: nominal curvature boundedness}
        \sup_{t\in[0,\infty)}
        \kappa_{\bm\xi}\!\left(\psi_{\bm\xi}(t;\bm p_0)\right)
        <\infty.
    \end{equation}
\end{assumption}

{To make sure the aligned integral curve remains collision-free after the deformation, the deformation should be confined within the free space, which requires the initial position to lie in $\operatorname{int}(\mathcal F)$.
To preserve convergence to the target limit set, we additionally require the initial position to belong to its region of attraction $\mathcal R_{\bm\xi}^{a}$.
We therefore restrict the admissible initial configurations in Definition~\ref{def:admissible initial set} as follows.
}

\begin{definition}
    \label{def:feasible initial set}
    An initial configuration $\bm g=(\bm p,\mathbf R)\in\mathcal G_a$ is called \emph{feasible}, if it belongs to the set
    \begin{equation*}\label{eq: feasible initial set}
        \mathcal G_f
        \triangleq
        \left\{
        \bm g=(\bm p,\mathbf R)\in\mathcal G_a
        \,\middle|\,
        \bm p\in\operatorname{int}(\mathcal F)\cap\mathcal R_{\bm\xi}^{a}
        \right\},
    \end{equation*}
    where the position lies in the intersection between the interior of the free space and the attraction region of target limit set.
\end{definition}

Denote the image of the nominal target limit set under DIVAL by $\mathcal L_{\bm\eta}^{+}\triangleq\Phi(\mathcal L_{\bm\xi}^{+})$.
The following theorem establishes collision avoidance and convergence to $\mathcal L_{\bm\eta}^{+}$ for feasible initial configurations when the active region is contained in the free space.

\begin{theorem}
\label{thm: safety and convergence}
{
Let $\bm\xi$ be a nominal vector field satisfying Assumptions~\ref{ass:nominal smoothness} and \ref{ass:nominal safety and convergence}.
Given an initial configuration $\bm g_0=(\bm p_0,\mathbf R_0)\in\mathcal G_f$, choose any active-region radius $0<r_{\mathcal A}\le\operatorname{dist}\!\left(\bm p_0,\partial\mathcal F\right)$ and any transition function $\lambda\in\Lambda^{r+1}(\bm p_0,r_{\mathcal A})$.}
Then, the closed-loop trajectory under the control laws \eqref{eq: v} and \eqref{eq: Omega} remains safe and converges to the deformed target limit set $\mathcal L_{\bm\eta}^{+}$, i.e.,
\begin{equation*}\label{eq: closed loop safety}
    \bm p(t;\bm g_0)\in\mathcal F,\quad \forall t\ge0,
\end{equation*}
and
\begin{equation*}\label{eq: closed loop convergence}
    \lim_{t\to\infty}
    \mathrm{dist}\left(\bm p(t;\bm g_0),\mathcal L_{\bm\eta}^{+}\right)=0.
\end{equation*}
\end{theorem}
\begin{proof}
    See Appendix~\ref{appendix:proof_safety_and_convergence}.
\end{proof}

Since $\Phi$ is the identity mapping for points outside the active region $\mathcal A$, the nominal target limit set remains unchanged whenever $r_{\mathcal A}\le\mathrm{dist}(\bm p_0,\mathcal L_{\bm\xi}^{+})$.
In this case, $\mathcal L_{\bm\eta}^{+}=\mathcal L_{\bm\xi}^{+}$, and the closed-loop system converges to the nominal target limit set:
\[
    \lim_{t\to\infty}
    \mathrm{dist}\left(\bm p(t;\bm g_0),\mathcal L_{\bm\xi}^{+}\right)=0.
\]

\subsubsection{Smoothness of Control Inputs}

\begin{figure}[h]
    \centering
    \begin{subfigure}[t]{0.45\columnwidth}
        \centering
        \includegraphics{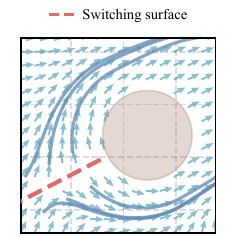}
        \caption{Vector field in \cite{Panagou2017Distributed}.}
        \label{fig: panagou switching surface}
    \end{subfigure}
    \hfill
    \begin{subfigure}[t]{0.45\columnwidth}
        \centering
        \includegraphics{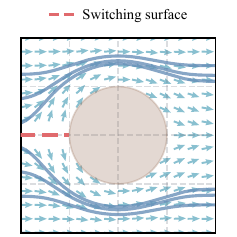}
        \caption{Vector field in \cite{Marchidan2020Collision}.}
    \end{subfigure}
    \caption{Examples of switching surfaces in obstacle-avoidance vector fields.}
    \label{fig:separatrix}
\end{figure}

Theorem~\ref{thm: safety and convergence} ensures the safety and convergence of the closed-loop system.
We next examine the smoothness of the control inputs along the closed-loop trajectory, with particular attention to positions where the vector field vanishes or is undefined.
In many obstacle-avoidance vector fields, the choice of obstacle-avoidance direction becomes non-unique at certain positions.
Some formulations assign a zero vector to these positions \cite{Panagou2017Distributed}, whereas others leave the vector field undefined there \cite{Marchidan2020Collision}.
In this paper, we refer to such sets of positions as switching surfaces, with examples shown in Figure~\ref{fig:separatrix}.

Encountering these switching surfaces would prevent a nominal integral curve from converging to the target. Under Assumptions~\ref{ass:nominal smoothness} and~\ref{ass:nominal safety and convergence}, nominal integral curves starting in the target's region of attraction \(\mathcal R_{\bm\xi}^{a}\) therefore avoid these surfaces.
However, under the vector-field-tracking paradigm, the robot's trajectory deviates from the nominal integral curve while the heading mismatch decreases over time.
During this deviation, the robot may encounter a switching surface that causes undefined or discontinuous control inputs.

DIVAL avoids this issue by removing the initial heading mismatch and ensuring that the robot follows the aligned integral curve throughout its motion, as established in Theorems~\ref{thm:aligned vf}~and~\ref{thm: trajectory relation}.
The diffeomorphism $\Phi$ deforms both the nominal integral curve and the switching surfaces of the nominal vector field.
When the initial position lies in $\mathcal R_{\bm\xi}^{a}$, the nominal integral curve avoids these surfaces.
Since $\Phi$ is one-to-one, the aligned integral curve likewise avoids the deformed switching surfaces, and so does the robot's trajectory.

With the trajectory avoiding the deformed switching surfaces, DIVAL guarantees control inputs of class $C^{r-1}$ for a nominal vector field of class $C^r$ as stated in the following theorem.

\begin{theorem}
\label{thm: smoothness}
{
Let $\bm\xi\in C^r(\mathcal D_{\bm\xi})$ with integer $r\ge1$ be a nominal vector field satisfying Assumptions~\ref{ass:nominal smoothness} and \ref{ass:nominal safety and convergence}.
Given an initial configuration $\bm g_0\in\mathcal G_f$ and an aligned vector field constructed under the conditions of Theorem~\ref{thm: safety and convergence}, the control inputs \eqref{eq: v} and \eqref{eq: Omega} are $C^{r-1}$ functions of the robot's configuration $\bm g(t;\bm g_0)$.
}
\end{theorem}
\begin{proof}
    Let $\bm g_0\in\mathcal G_f$.
    Theorem~\ref{thm:aligned vf} ensures that $\bm\eta\in C^r(\mathcal D_{\bm\eta})$, and the proof of Theorem~\ref{thm: safety and convergence} establishes that $\bm\eta$ remains non-singular along the closed-loop trajectory for all $t\ge0$.
    Since $\mathbf J_{\bm\eta}\in C^{r-1}$ and $k\in C^{r-1}$, the control laws \eqref{eq: v} and \eqref{eq: Omega} are $C^{r-1}$ functions of the configuration.
\end{proof}

\begin{remark}
    \label{remark: research gap}
    {Some works using the existing vector-field-tracking methods, such as \cite{qiao2024motion,yang2025finite,yao2023guiding,qiao2026curvature}, prove the convergence of the closed-loop system.}
    However, these results are limited to obstacle-free environments.
    In contrast, Theorems~\ref{thm: safety and convergence}~and~\ref{thm: smoothness} not only analyze DIVAL in obstacle environments but prove that the closed-loop trajectory remains safe, convergent, and admits continuous control inputs during convergence.
    These results extend existing vector-field-based planning by showing how the nominal vector field's various properties can be preserved in the closed-loop system.
\end{remark}

{
\label{page: bounded v}
We next show how the scaling factor $k$ accommodates the linear-velocity constraint $v = k \| \bm\eta \| \in [v_{\min}, v_{\max}]$ for different types of target.
When the target is an equilibrium where $\bm \eta = \bm 0$, the robot's speed must vanish when converging to it and hence $v_{\min}=0$ is required.
This lower bound is naturally accommodated by the vector field, since $\|\bm\eta\|\ge 0$ yields $v\ge v_{\min} = 0$ while the upper bound can be enforced by bounding $k$ from above at $v_{\max}/\|\bm\eta\|$.
For a limit-cycle target, a positive minimum speed $v_{\min}>0$ is needed to maintain motion along the limit cycle.
Both velocity bounds can be enforced by bounding $k$ to $[v_{\min}/\|\bm\eta\|,\,v_{\max}/\|\bm\eta\|]$.
These bounds remain finite because the limit cycle's region of attraction contains no equilibrium and hence $\bm\eta$ is nonzero along the trajectory converging to a compact limit cycle.}

{
\label{dis: omega limit}
When the above range of linear velocity includes zero, the angular velocity must vanish as the linear velocity approaches zero to exclude rotation in place as required in Section~\ref{sec:problem}.
Based on Assumption~\ref{ass:nominal safety and convergence}, the derivation in Appendix~\ref{appendix:curvature_bound} shows that the curvature $\kappa_{\bm\eta}$ along the aligned integral curve $\psi_{\bm \eta}(t;\bm p_0)$ remains finite under the deformation.
Therefore, with Theorem~\ref{thm: trajectory relation} ensuring that the robot moves along the aligned integral curve, substitution of the linear velocity \eqref{eq: v} and the curvature \eqref{eq: curvature bound split} into the angular velocity \eqref{eq: Omega} yields $\| \Omega \| = v \kappa_{\bm \eta} \to 0$ as $v \to 0$.

With the control laws derived and the conditions established for the robot to preserve the nominal vector field's properties, Algorithm~\ref{alg:dival} summarizes the resulting motion-planning procedure in two phases: DIVAL construction and continuous-time control input computation.}
\begin{algorithm}[H]
    \caption{DIVAL-Based Nonholonomic Motion Planning}
    \label{alg:dival}
    \small
    \begin{algorithmic}[1]
    \REQUIRE Nominal vector field $\bm\xi$ satisfying Assumptions~\ref{ass:nominal smoothness} and~\ref{ass:nominal safety and convergence};
    \item[] Initial configuration $\bm g_0\in\mathcal G_f$ as defined in Definition~\ref{def:feasible initial set}.
    \ENSURE Control inputs $(v,\bm\Omega)$.
    \item[] \hspace*{-\leftmargin}\textbf{DIVAL construction}
    \STATE Compute the rotation vector $\bm\Delta_*$ using \eqref{eq: Delta_*}.
    \STATE Select $r_{\mathcal A}$ and $\lambda$ satisfying the conditions of Theorem~\ref{thm: safety and convergence}.
    \STATE Construct the deformation map $\Phi$ using \eqref{eq: diffeomorphism 3D}.
    \item[] \hspace*{-\leftmargin}\textbf{Control input computation}
    \STATE Obtain the current configuration $\bm g(t)=(\bm p(t),\mathbf R(t))$.
    \STATE Evaluate the aligned vector field $\bm\eta(\bm p(t))$ using \eqref{eq: aligned vf}.
    \STATE Compute the control inputs $(v(t),\bm\Omega(t))$ using \eqref{eq: v} and \eqref{eq: Omega}.
    \end{algorithmic}
\end{algorithm}

\section{Simulation and Hardware Validation}
\label{sec:validation}

In this section, we first present design choices to improve DIVAL's implementation performance.
We then compare DIVAL with baselines in simulation through case studies and Monte Carlo analysis.
Finally, hardware experiments on a nonholonomic UGV further evaluate DIVAL under real-world conditions.

\subsection{Implementation of DIVAL}
\label{subsec:implementation}

{
The implementation follows the procedure summarized in Algorithm~\ref{alg:dival}.
For initialization, we select parameters of DIVAL to reduce the angular speeds required to follow the aligned integral curve.
During the robot's motion, we augment the control laws to compensate for disturbances in practice.

\subsubsection{Parameter Selection}
\label{subsubsec:param}

The initial configuration determines the deformation center $\bm p_0$ and the rotation vector $\bm\Delta_*$ that aligns the nominal vector field with the robot's initial heading, leaving the active-region radius $r_{\mathcal A}$ and transition function $\lambda$ to be selected.
Recall that the preceding analysis requires an active-region radius $0<r_{\mathcal A}\le\operatorname{dist}(\bm p_0,\partial\mathcal F)$ to keep the active region within the free space and an admissible transition function $\lambda\in\Lambda^{r+1}(\bm p_0,r_{\mathcal A})$ defined in Definition~\ref{def:admissible transition functions}.
Within these constraints, we first select the active-region radius based on the distance to the free-space boundary and then optimize the transition function over the resulting active region.

Under the control laws \eqref{eq: v} and \eqref{eq: Omega}, the robot moves along the aligned integral curve, so its angular speed and the curvature of this curve satisfy $\|\bm\Omega\|=v\kappa_{\bm\eta}$.
In Appendix~\ref{appendix:curvature_bound}, we derive the following relation between the aligned and nominal vector fields' curvatures
\begin{equation}
\label{eq: aligned curvature}
    \kappa_{\bm\eta}(\bm p) \le C_1(\bm q)\kappa_{\bm\xi}(\bm q)+C_2(\bm q),
\end{equation}
where $C_1(\bm q)$, $C_2(\bm q)$ are scalar functions of $\bm q$ defined in appendix, and $\bm q=\Phi^{-1}(\bm p)$.
On the right-hand side, the first term $C_1(\bm q)\kappa_{\bm\xi}(\bm q)$ quantifies the contribution of the nominal vector field's curvature $\kappa_{\bm \xi}$ under the deformation, and the second term $C_2(\bm q)$ quantifies the additional contribution induced by the spatial variation of deformation.

For a fixed initial heading mismatch, a fixed transition function, and a fixed value of $\bar{\bm q}/r_{\mathcal A}$, $C_1$ remains unchanged while $C_2$ scales as $r_{\mathcal A}^{-1}$.
This motivates selecting the largest radius allowed by the safety condition in Theorem~\ref{thm: safety and convergence} to reduce the second term in \eqref{eq: aligned curvature} and lower angular-speed demands
\begin{equation}
\label{eq: max r_A}
    r_{\mathcal A}
    =d_{\mathcal F}(\bm p_0)
    \triangleq
    \operatorname{dist}(\bm p_0,\partial\mathcal F),
\end{equation}
where $\mathcal F$ is the free space and $d_{\mathcal F}$ is the distance from the initial position $\bm p_0$ to its boundary.

With the active-region radius fixed, we proceed to optimize the transition function.
For this purpose, we parameterize its profile using a degree-five B-spline \cite[Appendix B]{biagiotti2008trajectory} with 12 coefficients $\bm c = [c_0,c_1,\dots,c_{11}]$ as follows
\begin{equation}
\label{eq: spline transition}
\begin{aligned}
    \lambda_{\bm c}(\bm q)
    &=
    \begin{cases}
    \displaystyle\sum_{i=0}^{11}c_i B_{i}(s(\bm q)),
    &0\le s(\bm q)<1,\\
    0,&s(\bm q)\ge1,
    \end{cases}\\
    s(\bm q)&\triangleq\frac{\|\bm q-\bm p_0\|}{r_{\mathcal A}}.
\end{aligned}
\end{equation}
Here, $B_{i}$ are B-spline basis functions with equally spaced interior knots, and the coefficients satisfy
\begin{equation}
\label{eq: spline coefficient constraint}
\bm c\in
\mathcal C \triangleq
\big\{\bm c\in[0,1]^{12} \mid
c_{0,1}=1,
c_{9,10,11}=0
\big\}.
\end{equation}
Specifically, each endpoint knot of $B_i$ is repeated six times so that the coefficient constraints enforce unit value and zero first derivative at $s=0$, and zero value, first derivative, and second derivative at $s=1$.
Together with the coefficient bounds, these conditions ensure that the transition function \eqref{eq: spline transition} belongs to $\Lambda^2(\bm p_0,r_{\mathcal A})$.

We optimize the B-spline coefficients to reduce both the overall and peak curvature of the aligned integral curve starting from $\bm p_0$.
For each $\bm c$, let $L_{\bm c}$ be the length of the aligned integral curve's segment within $\mathcal A$, and denote its curvature at arc length $\ell$ by $\kappa_{\bm c}(\ell)$.
We define the objective as the sum of the root-mean-square (RMS) and peak curvatures given by
\begin{equation}
\label{eq: lbd opt}
    J(\bm c)\triangleq
    \left(\frac{1}{L_{\bm c}}\int_0^{L_{\bm c}}
    \kappa_{\bm c}(\ell)^2\,d\ell\right)^{1/2}
    +\max_{\ell\in[0,L_{\bm c}]}\kappa_{\bm c}(\ell).
\end{equation}
By minimizing the above objective subject to \eqref{eq: spline coefficient constraint}, we obtain the coefficients $\bm c^\star \in \mathcal{C}$ and use the optimized transition function $\lambda^\star=\lambda_{\bm c^\star}$ throughout the robot's motion.
}

\begin{figure*}[!b]
    \centering
    \begin{subfigure}[t]{0.30\textwidth}
        \centering
        \includegraphics[width=\linewidth]{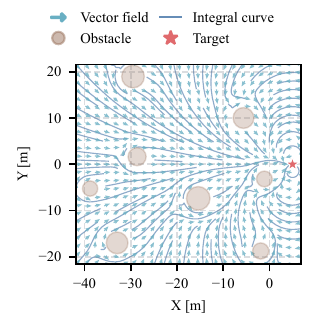}
        \caption{ARVF.}
        \label{fig:mc_layout_arvf}
    \end{subfigure}
    \hfill
    \begin{subfigure}[t]{0.30\textwidth}
        \centering
        \includegraphics[width=\linewidth]{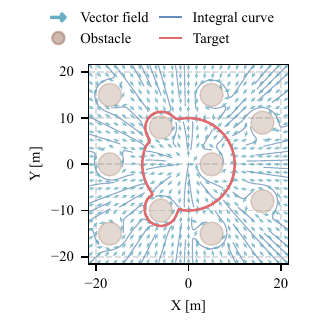}
        \caption{CAPF-VF.}
        \label{fig:mc_layout_vfcapf}
    \end{subfigure}
    \hfill
    \begin{subfigure}[t]{0.30\textwidth}
        \centering
        \includegraphics[width=\linewidth]{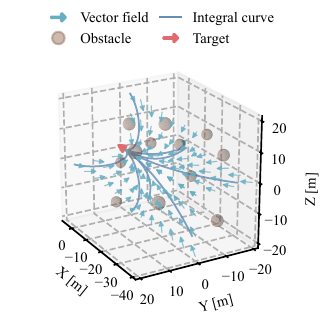}
        \caption{NVF.}
        \label{fig:mc_layout_nvf}
    \end{subfigure}
    \caption{
        Nominal vector field and obstacle layout for numerical simulations.
        The ARVF and CAPF-VF are evaluated in 2D while NVF is in 3D.
        The target positive limit sets of these vector fields are a target point, a target limit cycle, and a target point with specified approaching direction, respectively.
        }
    \label{fig:mc_layout_fields}
\end{figure*}

{
\subsubsection{Robust Control Law}
\label{subsubsec: robust}
Disturbances in practice, such as measurement errors, actuator imperfections and terrain variations, will affect the accuracy of heading alignment and cause the robot to deviate from the motion described by the kinematic model.
To compensate for possible deviations, we augment the derived control laws with feedback based on the difference between the actual configuration and a reference configuration defined as follows.
The reference configuration $\bm g(t)=(\bm p(t),\mathbf R(t))$ with the heading $\bm h = \mathbf{R} \bm e_1$ is generated by integrating the kinematics \eqref{eq:kinematics} under the feedforward control laws \eqref{eq: v} and \eqref{eq: Omega}.
Let $\bm g_{\mathrm{act}}=(\bm p_{\mathrm{act}},\mathbf R_{\mathrm{act}})$ and $\bm h_{\mathrm{act}}=\mathbf R_{\mathrm{act}}\bm e_1$ denote the actual configuration and heading, respectively.
The longitudinal, lateral, and heading errors are defined as
\begin{equation}
\label{eq: practical errors}
\begin{aligned}
    e_{\parallel}
    &\triangleq
    \langle\bm h_{\mathrm{act}},\bm p-\bm p_{\mathrm{act}}\rangle,\\
    \bm e_{\perp}
    &\triangleq
    (\mathbf I-\bm h_{\mathrm{act}}\bm h_{\mathrm{act}}^\intercal)(\bm p-\bm p_{\mathrm{act}}),\\
    \bm e_h
    &\triangleq
    \bm h_{\mathrm{act}}\times\bm h.
\end{aligned}
\end{equation}
Here, $e_{\parallel}$ is the projection of the position error $\bm p-\bm p_{\mathrm{act}}$ onto the actual heading $\bm h_{\mathrm{act}}$, and $\bm e_{\perp}$ is the projection of $\bm p - \bm p_{\mathrm{act}}$ onto the plane orthogonal to $\bm h_{\mathrm{act}}$.
The heading error $\bm e_h$ has magnitude equal to the sine of the angle between the actual and reference headings, and its direction gives the rotation axis from $\bm h_{\mathrm{act}}$ toward $\bm h$.

Augmenting the feedforward terms with the above error feedback gives the following robust control law
\begin{subequations}
\label{eq: commanded inputs}
\begin{gather}
v_{\mathrm{fb}}
=
v(\bm g)\langle\bm h_{\mathrm{act}},\bm h\rangle
+k_v e_{\parallel},
\label{eq: commanded v}\\
\begin{aligned}
    \bm\Omega_{\mathrm{fb}}
    =
    \bm\Omega_\mathrm{ff}
    +\mathbf R_{\mathrm{act}}^\intercal v(\bm g)
    \left(k_h\bm e_h + k_p \bm h_{\mathrm{act}}\times\bm e_{\perp}\right),
\end{aligned}
\label{eq: commanded Omega}
\end{gather}
\end{subequations}
where $k_v$, $k_h$, $k_p$ are constant positive feedback gains, and $\bm\Omega_\mathrm{ff} = \mathbf R_{\mathrm{act}}^\intercal\mathbf R\bm\Omega(\bm g)$ is the feedforward angular velocity expressed in the actual robot's body frame. The structure of \eqref{eq: commanded inputs} extends the tracking law of \cite{kanayama1990stable} to a configuration-dependent feedforward velocity $v(\bm g)$ and the reference angular velocity $\bm\Omega(\bm g)$.
When the actual configuration matches the reference, i.e., $\bm g_\mathrm{act} = \bm g$, the robust control law \eqref{eq: commanded inputs} reduces to \eqref{eq: v} and \eqref{eq: Omega}.
For planar motion, the aligned vector field can be constructed using \eqref{eq: diffeomorphism 2D}, and the reference configuration is propagated through \eqref{eq:kinematics 2d} using \eqref{eq: v 2d} and \eqref{eq: omega 2d}. The third component of \eqref{eq: commanded Omega} gives the angular velocity input.

\begin{proposition}
\label{prop:tracking error bound}
Consider the reference position $\bm p(t)$ and reference heading $\bm h(t)$ generated by \eqref{eq: v} and \eqref{eq: Omega}, and let
$d(t)\triangleq\|\bm p_{\mathrm{act}}(t)-\bm p(t)\|^2+\frac{2}{k_p}\left(1-\langle\bm h_{\mathrm{act}}(t),\bm h(t)\rangle\right)$.
Under the robust control law \eqref{eq: commanded inputs}, the reference $(\bm p(t),\bm h(t))$ is locally stable.
Moreover, if a perturbation acts on $[0,t_{\mathrm p})$ and vanishes for $t\ge t_{\mathrm p}$, the position error remains bounded by
\begin{equation}
\label{eq: tracking error bound}
\|\bm p_{\mathrm{act}}(t)-\bm p(t)\|\le\sqrt{d(t_{\mathrm p})}
\triangleq\varepsilon_{\mathrm p},\quad t\ge t_{\mathrm p}.
\end{equation}
\end{proposition}
\begin{proof}
Let
$$V\triangleq\frac{k_p}{2}d=\frac{k_p}{2}\|\bm p-\bm p_{\mathrm{act}}\|^2+1-\langle\bm h_{\mathrm{act}},\bm h\rangle,$$
which is positive definite with respect to $(\bm p,\bm h)$ and vanishes only at $\bm p_{\mathrm{act}}=\bm p$, $\bm h_{\mathrm{act}}=\bm h$. Along the closed-loop system, \eqref{eq: practical errors} and \eqref{eq: commanded inputs} give
\begin{equation}
\label{eq: Lyapunov derivative}
\begin{aligned}
\dot V
&=
k_p\left(v(\bm g)\left\langle\bm e_{\perp},\bm h\right\rangle-k_v e_{\parallel}^2\right)\\
&\quad-v(\bm g)\left(k_h\|\bm e_h\|^2+k_p\left\langle\bm e_{\perp},\bm h\right\rangle\right)\\
&=
-k_pk_v e_{\parallel}^2-k_hv(\bm g)\|\bm e_h\|^2\le0,
\end{aligned}
\end{equation}
where the mixed terms cancel through the lateral feedback $k_p\bm h_{\mathrm{act}}\times\bm e_{\perp}$ in \eqref{eq: commanded Omega}. Hence $V$ is nonincreasing, so the Lyapunov method proves the stability claim; and since \eqref{eq: Lyapunov derivative} holds for $t\ge t_{\mathrm p}$, we have $d(t)\le d(t_{\mathrm p})$, which together with $\|\bm p_{\mathrm{act}}-\bm p\|^2\le d$ gives \eqref{eq: tracking error bound}.
\end{proof}

\begin{remark}
    \label{remark: perturbated properties}
    In the case of perturbations, the safety of the actual trajectory $\bm p_\mathrm{act}$ can be established by inflating each obstacle by a margin greater than $\varepsilon_{\mathrm p}$ and then by constructing the nominal vector field and DIVAL in the resulting free space.
    Regarding the convergence, since the reference trajectory $\bm p$ converges to the target set, the distance from $\bm p_\mathrm{act}$ to that set is bounded by $\varepsilon_{\mathrm p}$ as time goes to infinity.
    Moreover, the feedback terms are smooth functions of the reference and actual configurations, so the robust control inputs \eqref{eq: commanded inputs} retain the $C^{r-1}$ smoothness of the feedforward terms \eqref{eq: v} and \eqref{eq: Omega}.
\end{remark}
}

\subsection{Numerical Simulations}
\label{subsec:simulation}

This subsection presents two sets of simulation results: case studies illustrating safety, convergence, control-input continuity and robustness, and Monte Carlo simulations assessing performance across initial configurations and the effects of parameter selection.
We first introduce the setup shared by both.

\subsubsection{Setup}

We consider three vector-field-tracking baselines: the attractive and repulsive vector field (ARVF) \cite[Eq.~17]{Panagou2017Distributed}, the collision avoidance and path following vector field (CAPF-VF) \cite[Eq.~7]{yao2022guiding}, and the navigation vector field (NVF) \cite[Eq.~50]{he2025novel}, which are visualized in Figure~\ref{fig:mc_layout_fields}.
Their angular velocities in \cite[Eq.~18(b)]{Panagou2017Distributed}, \cite[Sec.~VI.D]{yao2022guiding}, and \cite[Eq.~32]{he2025novel} use the heading error feedback with gain $k_{\bm \Omega}>0$ to gradually rotate the robot's heading toward the vector field.
For each baseline, we apply DIVAL to the same nominal vector field.

\begin{table}[!t]
    \centering
    \caption{
    Initial heading mismatch magnitudes for the case studies, in radians.}
    \label{tab:case studies}
    \renewcommand{\arraystretch}{1.12}
    \setlength{\tabcolsep}{3pt}
    \begin{tabular}{c c c c}
        \toprule
         & \textbf{CAPF-VF} & \textbf{NVF} & \textbf{ARVF} \\
         & $\lvert\delta_*\rvert$ & $\|\bm\Delta_*\|$ & $\lvert\delta_*\rvert$ \\
        \midrule
        Set 1 & 1.9168 & 1.0220 & 1.3981 \\
        Set 2 & 2.3295 & 2.2742 & 1.2652 \\
        Set 3 & 1.4464 & 1.6239 & 0.3235 \\
        \bottomrule
    \end{tabular}
\end{table}

\begin{figure}[!b]
    \centering
    \begin{subfigure}{0.9\columnwidth}
        \centering
        \includegraphics{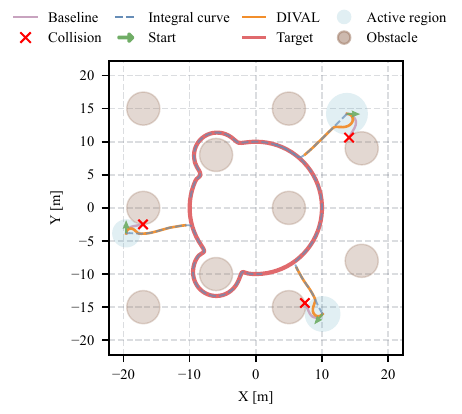}
        \caption{CAPF-VF, red curve denoting the target limit cycle.}
    \end{subfigure}
    \begin{subfigure}{0.9\columnwidth}
        \centering
        \includegraphics{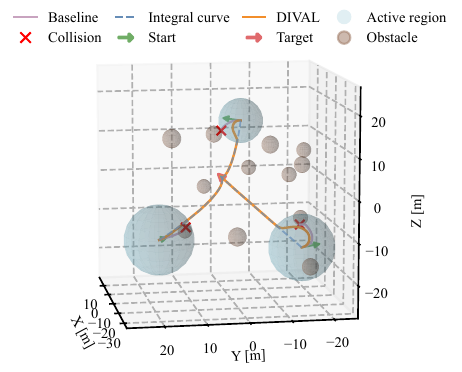}
        \caption{NVF, red arrow denoting the target position and heading.}
    \end{subfigure}
    \caption{Simulation results based on the CAPF-VF and the NVF, where the baselines with gain $k_{\bm \Omega} = 0.1$ encounter collisions, while the DIVAL variants demonstrate safe convergence.}
    \label{fig:case study safe}
\end{figure}

The linear-velocity commands of the baselines follow two representative strategies.
When tracking CAPF-VF, the controller in \cite[Sec.~VI.D]{yao2022guiding} uses a constant speed.
When tracking ARVF and NVF, the robot decelerates as it approaches the target position under the controllers in \cite[Eq.~18(a)]{Panagou2017Distributed} and \cite[Eq.~38]{he2025novel}.
Let $v_{\mathrm{base}}(\bm p)$ denote the baseline's linear velocity at position $\bm p$.
Since \eqref{eq: v} gives the linear velocity of DIVAL by $v(\bm p)=k(\bm p)\|\bm\eta(\bm p)\|$, we choose
\begin{equation*}
    k(\bm p)=\frac{v_{\mathrm{base}}(\bm p)}{\|\bm{\eta}(\bm p)\|}
\end{equation*}
at non-singular positions of $\bm\eta$, so that DIVAL's linear velocity equals the baseline command for the same position.
To compensate for discretization errors when implementing the continuous-time control laws in simulation, we use their robust version in \eqref{eq: commanded inputs} with gains $(k_v,k_h,k_p)=(1,1,1)$ in 2D and $(1,3,3)$ in 3D.

\begin{figure}[!t]
    \centering
    \begin{subfigure}{0.95\columnwidth}
        \centering
        \includegraphics[width=\linewidth]{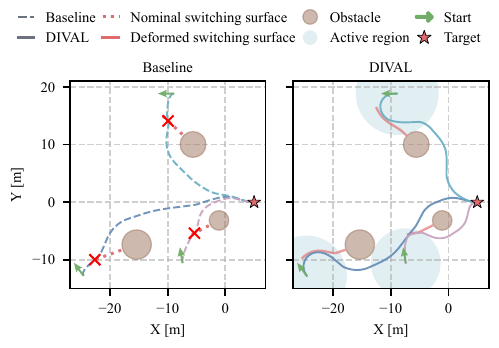}
        \caption{Closed-loop trajectories. Red crosses denote the intersection between the trajectories and the switching surfaces.}
        \label{fig: switching trajectories}
    \end{subfigure}

    \begin{subfigure}{0.95\columnwidth}
        \centering
        \includegraphics[width=\linewidth]{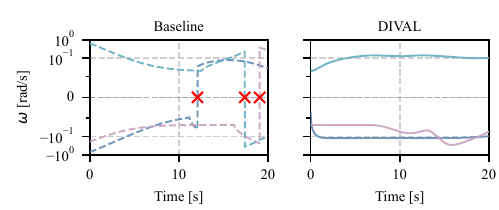}
        \caption{Angular velocity, red crosses denote discontinuity.}
        \label{fig: switching omega}
    \end{subfigure}
    \caption{Simulation results based on ARVF. For visual clarity, subfigure~(\subref{fig: switching omega}) shows the first 20 seconds. The baseline angular velocities with gain $k_{\bm \Omega}=0.5$ jump when the trajectories cross the switching surfaces, whereas the DIVAL angular velocities remain continuous over the displayed interval.}
    \label{fig:case study continuity}
\end{figure}

\subsubsection{Case Studies}

For the case studies, the active-region radius and transition function are selected according to \eqref{eq: max r_A} and \eqref{eq: lbd opt}.

The CAPF-VF and NVF cases examine safety and convergence.
According to \cite[Thm.~1]{yao2022guiding} \cite[Thm.~3]{he2025novel}, the nominal integral curves of the CAPF-VF and NVF avoid obstacles and converge to the target limit cycle and position, respectively.
However, with the initial heading mismatches listed in Table~\ref{tab:case studies}, the baseline closed-loop trajectories collide with obstacles due to the deviation from the nominal integral curves, as shown in Figure~\ref{fig:case study safe}.
In contrast, the closed-loop trajectories of DIVAL follow the aligned integral curves and reach the targets without collision, validating the safety and convergence results in Theorem~\ref{thm: safety and convergence}.

The ARVF cases examine control-input continuity.
In \cite[Fig.~4]{Panagou2017Distributed}, the ARVF is constructed as a piecewise $C^1$ vector field, except on the switching surfaces illustrated by red dotted lines in Figure~\ref{fig:case study continuity}(\subref{fig: switching trajectories}).
The baseline trajectories cross these switching surfaces, producing the angular-velocity jumps shown in Figure~\ref{fig:case study continuity}(\subref{fig: switching omega}).
As discussed in Section~\ref{subsec:guarantees}, the aligned integral curves avoid the deformed switching surfaces in Figure~\ref{fig:case study continuity}(\subref{fig: switching trajectories}), since the nominal integral curves do not intersect the switching surfaces as shown in Figure~\ref{fig:separatrix}(\subref{fig: panagou switching surface}).
By following the aligned integral curves, the angular-velocity commands of DIVAL remain continuous, consistent with Theorem~\ref{thm: smoothness}.

\begin{figure}[!htb]
    \centering
    \begin{subfigure}[t]{0.45\columnwidth}
        \centering
        \includegraphics[width=\linewidth]{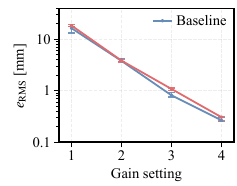}
        \caption{Error to the limit cycle}
        \label{fig:robustness_orbit}
    \end{subfigure}
    \hspace{0.04\columnwidth}
    \begin{subfigure}[t]{0.45\columnwidth}
        \centering
        \includegraphics[width=\linewidth]{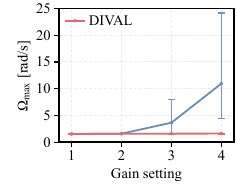}
        \caption{Peak angular velocity}
        \label{fig:robustness_omega}
    \end{subfigure}
    \caption{Robustness comparison under persistent input disturbances. Markers and error bars show medians and interquartile ranges across ten trials. Gain settings 1--4 correspond to $k_{\bm \Omega}=0.2,1,5,15$ for the baseline and $(k_v,k_h,k_p)=(1,1,1),\allowbreak(1,5,5),\allowbreak(1,25,25),\allowbreak(1,30,200)$ for DIVAL, respectively.}
    \label{fig:robustness_metrics}
\end{figure}

{In addition to control-input continuity, we compare the robustness of the baseline's attitude tracking controller and DIVAL's controller in \eqref{eq: commanded inputs} under persistent input disturbances.
The disturbances include additive deviations with standard deviations of \SI{0.005}{m/s} and \SI{0.01}{rad/s} for linear and angular velocity and multiplicative deviations with standard deviations of $1\%$ and $3\%$ for linear and angular velocity, respectively.
Figure~\ref{fig:robustness_metrics} reports the RMS distance error $e_{\mathrm{RMS}}$ relative to the stable limit cycle of CAPF-VF after $t=$\SI{100}{s}, and the peak commanded angular velocity $\Omega_{\max}$ over 10 trials with randomly sampled initial configurations.
Both methods achieve comparable tracking errors that decrease as the feedback gains increase.
However, this improvement increases the baseline's peak angular velocity to approximately seven times of its value at the lowest gain setting, whereas DIVAL's peak remains nearly unchanged.
This is because the magnitude of the initial heading mismatch varies from zero to $\pi$, and multiplying it by a larger feedback gain usually produces larger control inputs for the baseline.
In contrast, DIVAL eliminates the initial heading mismatch, so its feedback gains act only on small tracking errors induced by disturbances, making the control inputs less sensitive to gain increases.}

\subsubsection{Monte Carlo Simulation}
\label{subsubsec: MC}

\begin{table*}[p]
    \centering
    \caption{Monte Carlo results for selected feedback gains of the vector-field-tracking baselines and the DIVAL variants.
    $P_s$ is the empirical success rate.
    $\bar e_{\bm h}$, $\Omega_{\mathrm{RMS}}$, $\Omega_{\max}$, and $\alpha_{\max}$ are medians of heading error, RMS angular velocity, peak angular velocity, and peak angular acceleration across trials, in rad, rad/s, rad/s, and rad/s$^2$, respectively.
    Values are rounded to four decimal places.
    Arrows indicate preferred directions, and bold values mark the best displayed results separately for the baseline and DIVAL within each vector field.
    DIVAL settings are specified as $(r_{\mathcal A},\lambda)$, where $r_{\mathcal A}$ is the active-region radius and $\lambda$ is the transition function.
    }
    \label{tab:comparison_raw}
    \renewcommand{\arraystretch}{1.23}
    \setlength{\tabcolsep}{3pt}
    \setlength{\arrayrulewidth}{0.3pt}
    \newcommand{\mcRulePadTop}{\rule{0pt}{\dimexpr\arraystretch\ht\strutbox+1.5pt\relax}}
    \newcommand{\mcRulePadBottom}{\rule[-\dimexpr\arraystretch\dp\strutbox+1.5pt\relax]{0pt}{\dimexpr\arraystretch\dp\strutbox+1.5pt\relax}}
    \begin{tabularx}{0.86\textwidth}{@{}c|c|cCCCCC@{}}
    \specialrule{0.9pt}{0pt}{0pt}
    \mcRulePadTop\mcRulePadBottom\textbf{Vector field} & \textbf{Method} & \textbf{Setting} & $P_s$ $\uparrow$ & $\bar{e}_{\bm{h}}$ $\downarrow$ & $\Omega_{\mathrm{RMS}}$ $\downarrow$ & {$\Omega_{\max}$ $\downarrow$} & {$\alpha_{\max}$ $\downarrow$} \\
    \specialrule{0.55pt}{0pt}{0pt}
    \multirow{7}{*}{ARVF}
        & \multirow{4}{*}{Baseline} & \mcRulePadTop$k_{\bm \Omega}=0.02$ & 0.7170 & 0.5535 & \textbf{0.0401} & {\textbf{0.1102}} & {\textbf{4.2696}} \\
        & & $k_{\bm \Omega}=0.2$ & 0.9610 & 0.0765 & 0.0665 & {0.3606} & {4.4232} \\
        & & $k_{\bm \Omega}=1$ & 0.9920 & 0.0154 & 0.1216 & {1.5778} & {4.4586} \\
        & & \mcRulePadBottom$k_{\bm \Omega}=5$ & \textbf{1.0000} & \textbf{0.0031} & 0.2564 & {7.7860} & {37.9872} \\
    \cline{2-8}
        & \multirow{3}{*}{{DIVAL}} & {\mcRulePadTop$(d_{\mathcal F}/2,\lambda_0)$} & {\textbf{1.0000}} & {\textbf{0.0000}} & {0.1480} & {2.1136} & {9.8193} \\
        & & {$(d_{\mathcal F},\lambda_0)$} & {\textbf{1.0000}} & {\textbf{0.0000}} & {0.1036} & {1.0868} & {5.4308} \\
        & & {\mcRulePadBottom$(d_{\mathcal F},\lambda^\star)$} & {\textbf{1.0000}} & {\textbf{0.0000}} & {\textbf{0.0747}} & {\textbf{0.2006}} & {\textbf{4.5223}} \\
    \specialrule{0.55pt}{0pt}{0pt}
    \multirow{7}{*}{CAPF-VF}
        & \multirow{4}{*}{Baseline} & \mcRulePadTop$k_{\bm \Omega}=0.02$ & 0.6480 & 0.5107 & \textbf{0.1273} & {0.7799} & {0.7763} \\
        & & $k_{\bm \Omega}=0.2$ & 0.9310 & 0.2240 & 0.1694 & {\textbf{0.6103}} & {\textbf{0.4550}} \\
        & & $k_{\bm \Omega}=1$ & 0.9930 & 0.0523 & 0.2612 & {1.6034} & {1.6263} \\
        & & \mcRulePadBottom$k_{\bm \Omega}=5$ & \textbf{1.0000} & \textbf{0.0110} & 0.5260 & {7.6830} & {37.6614} \\
    \cline{2-8}
        & \multirow{3}{*}{{DIVAL}} & {\mcRulePadTop$(d_{\mathcal F}/2,\lambda_0)$} & {\textbf{1.0000}} & {\textbf{0.0000}} & {0.3648} & {3.0087} & {19.1615} \\
        & & {$(d_{\mathcal F},\lambda_0)$} & {\textbf{1.0000}} & {\textbf{0.0000}} & {0.2538} & {1.6032} & {5.5069} \\
        & & {\mcRulePadBottom$(d_{\mathcal F},\lambda^\star)$} & {\textbf{1.0000}} & {\textbf{0.0000}} & {\textbf{0.1978}} & {\textbf{0.6076}} & {\textbf{1.4848}} \\
    \specialrule{0.55pt}{0pt}{0pt}
    \multirow{7}{*}{NVF}
        & \multirow{4}{*}{Baseline} & \mcRulePadTop$k_{\bm \Omega}=0.1$ & 0.8590 & 0.4305 & 2.2151 & {499.5076} & {2092.8046} \\
        & & $k_{\bm \Omega}=0.5$ & 0.9910 & 0.1013 & \textbf{0.2176} & {\textbf{1.2139}} & {\textbf{0.5790}} \\
        & & $k_{\bm \Omega}=2$ & 0.9980 & 0.0259 & 0.4241 & {4.7628} & {9.3303} \\
        & & \mcRulePadBottom$k_{\bm \Omega}=10$ & \textbf{0.9990} & \textbf{0.0053} & 0.9342 & {23.6203} & {224.5494} \\
    \cline{2-8}
        & \multirow{3}{*}{{DIVAL}} & {\mcRulePadTop$(d_{\mathcal F}/2,\lambda_0)$} & {\textbf{1.0000}} & {\textbf{0.0000}} & {0.2485} & {3.7704} & {31.2869} \\
        & & {$(d_{\mathcal F},\lambda_0)$} & {\textbf{1.0000}} & {\textbf{0.0000}} & {0.1701} & {1.8306} & {7.5675} \\
        & & {\mcRulePadBottom$(d_{\mathcal F},\lambda^\star)$} & {\textbf{1.0000}} & {\textbf{0.0000}} & {\textbf{0.1207}} & {\textbf{0.3465}} & {\textbf{1.5340}} \\
    \specialrule{0.9pt}{0pt}{0pt}
    \end{tabularx}
    \par\vspace{16pt}
    \captionsetup{type=figure}
    \centering
    \includegraphics[width=0.95\textwidth]{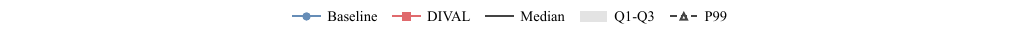}
    \begin{subfigure}[t]{0.32\textwidth}
        \centering
        \includegraphics{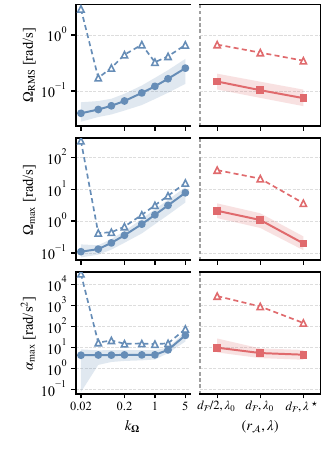}
        \caption{ARVF.}
        \label{fig:comparison_rms_arvf}
    \end{subfigure}
    \hfill
    \begin{subfigure}[t]{0.32\textwidth}
        \centering
        \includegraphics{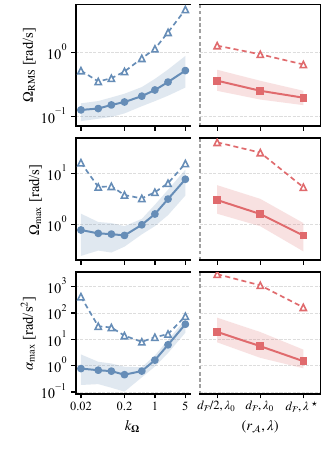}
        \caption{CAPF-VF.}
        \label{fig:comparison_rms_vf_capf}
    \end{subfigure}
    \hfill
    \begin{subfigure}[t]{0.32\textwidth}
        \centering
        \includegraphics{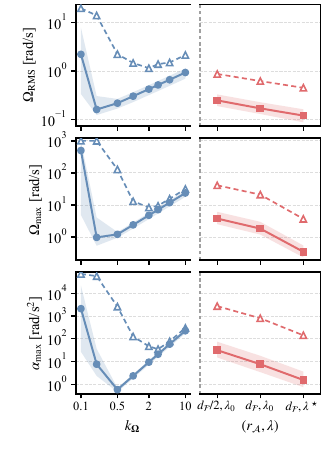}
        \caption{NVF.}
        \label{fig:comparison_rms_nvf}
    \end{subfigure}
    \caption{{Control effort of the vector-field-tracking baselines and DIVAL variants. Rows, from top to bottom, show angular-velocity RMS, peak angular velocity, and peak angular acceleration. Solid curves show medians, shaded bands indicate interquartile ranges, and dashed curves show the 99th percentiles across all 1000 trials. The vertical axes are logarithmic.}}
    \label{fig:comparison_rms}
\end{table*}

The Monte Carlo study examines how each baseline's feedback gain and the DIVAL parameter choices affect performance.
All methods are evaluated using the same 1000 randomly sampled initial configurations in the free space of each nominal vector field.
{
The vector-field-tracking baselines are evaluated at eight positive values of the feedback gain $k_{\bm \Omega}$.
For DIVAL, we construct three variants to examine the effects of increasing the active-region radius and optimizing the transition function.
We start with half of the maximum active-region radius $d_\mathcal{F}$ in \eqref{eq: max r_A} and use the fixed bump function
\begin{equation}
    \label{eq:transition_weight_exp}
    \begin{aligned}
    \lambda_0(\bm q)&=
    \begin{cases}
        \exp\!\left(-\dfrac{s(\bm q)^2}{1-s(\bm q)^2}\right), & 0\le s(\bm q)<1,\\
        0, & s(\bm q)\ge1,
    \end{cases}\\
    s(\bm q)&=\frac{\|\bm q-\bm p_0\|}{r_{\mathcal A}},
    \end{aligned}
\end{equation}
which provides a simple choice of transition function in $\Lambda^\infty(\bm p_0,r_{\mathcal A})$ and gives the first variant, denoted by $(r_{\mathcal A},\lambda)=(d_{\mathcal F}/2,\lambda_0)$.
We then increase the radius to the maximum while retaining the same bump profile, yielding the second variant $(d_{\mathcal F},\lambda_0)$.
Finally, we optimize the transition function by minimizing \eqref{eq: lbd opt} at the same radius, yielding the third variant $(d_{\mathcal F},\lambda^\star)$, which implements the full parameter selection strategy.
}

The following metrics are used to evaluate all baseline settings and DIVAL variants.
A Monte Carlo trial succeeds if the robot remains collision-free and reaches the target within prescribed tolerances before the simulation time limit.
The success rate $P_s$ is defined as the fraction of successful trials among all 1000 initial configurations and evaluates safety and convergence of the candidates.
For tracking accuracy, the heading error $\bar e_{\bm h}$ measures the angle between the robot heading and the vector-field direction.
This direction is evaluated at the robot's position using the nominal vector field for the baselines and the aligned vector field for DIVAL.
{Control effort is evaluated using three metrics: RMS angular velocity, peak angular velocity, and peak angular acceleration, denoted by $\Omega_{\mathrm{RMS}}$, $\Omega_{\max}$, and $\alpha_{\max}$, respectively.
The first two quantify the magnitude of the control input, whereas the third quantifies its variation.}
For each trial, we compute the mean heading error and the three control-effort metrics over its duration, then summarize these values across all trials, including failed trials.
Table~\ref{tab:comparison_raw} reports the results for four representative gains per baseline and all three DIVAL settings.
Figure~\ref{fig:comparison_rms} presents the control-effort distributions over the complete gain sweep and the three DIVAL settings.

Table~\ref{tab:comparison_raw} shows that the baseline controllers trade control effort for success rate and tracking accuracy as the gain increases.
For ARVF and CAPF-VF, the lowest gains yield small median control effort but success rates of only 0.7170 and 0.6480, respectively, with median heading errors exceeding $0.5\,\mathrm{rad}$.
For NVF, increasing the gain from 0.1 to 0.5 simultaneously improves all metrics, and further increases in gain reduce heading errors with a slightly higher success rate at the cost of significantly higher control effort.
However, even at the highest tested gain, $k_{\bm \Omega}=10$, the NVF baseline still fails in one trial, illustrating that increasing the gain only improves the success rate empirically rather than ensuring success.

In contrast, the DIVAL variants succeed in all trials and achieve near-zero heading errors while requiring lower control effort than the highest-gain baseline in each scene.
The observed success rates and heading accuracy are consistent with the theoretical guarantees of DIVAL.
According to \cite[Thm.~4]{Panagou2017Distributed}, \cite[Thm.~1]{yao2022guiding}, \cite[Thm.~1]{he2025novel}, the three nominal vector fields are non-singular and converge to their respective target limit sets from almost every initial position in the free space.
{Therefore, the initial configurations excluded by the feasible set $\mathcal G_f$ in Definition~\ref{def:feasible initial set} form measure-zero sets, and initial configurations randomly sampled from the free space fall in $\mathcal G_f$ with probability one.
For these configurations, Theorem~\ref{thm: safety and convergence} ensures success, while Theorems~\ref{thm:aligned vf} and~\ref{thm: trajectory relation} result in near-zero heading errors up to the numerical error.}

{
The proposed parameter-selection strategy reduces control effort while maintaining success rate and heading accuracy.
With transition function fixed as $\lambda_0$ in \eqref{eq:transition_weight_exp}, increasing the active-region radius from $d_{\mathcal F}/2$ to $d_{\mathcal F}$ reduces the medians of $\Omega_{\mathrm{RMS}}$, $\Omega_{\max}$, and $\alpha_{\max}$ by 30.0\%--31.6\%, 46.7\%--51.4\%, and 44.7\%--75.8\%, respectively.
Further optimization of the transition function at the maximum radius reduces these medians by 22.0\%--29.0\%, 62.1\%--81.5\%, and 16.7\%--79.7\%, respectively.
Compared with the highest-gain baselines achieving comparable success rates, the optimized DIVAL variant reduces the medians of $\Omega_{\mathrm{RMS}}$, $\Omega_{\max}$, and $\alpha_{\max}$ by 62.4\%--87.1\%, 92.1\%--98.5\%, and 88.1\%--99.3\%, respectively, across the three nominal vector fields.
}

\newsavebox{\hwsetupbox}
\setbox\hwsetupbox=\vbox{\setlength{\intextsep}{6pt}\begin{table}[H]
    \centering
    \footnotesize
    \caption{Experiment setup. $T_i$ denotes the $i$-th target position $\bm p_d$. $S_j$ denotes the $j$-th initial configuration, including the initial position $\bm p_0$ and initial heading mismatch $\delta_*$. Positions are in meters and heading mismatches are in radians.}
    \label{tab:exp_setup}
    \renewcommand{\arraystretch}{1.12}
    \renewcommand{\cellgape}{\Gape[1.5pt][1.5pt]}
    \setlength{\tabcolsep}{1.5pt}
    \setlength{\extrarowheight}{1pt}
    \setlength{\aboverulesep}{0pt}
    \setlength{\belowrulesep}{0pt}
    \begin{tabular*}{\columnwidth}{@{\extracolsep{\fill}}c|cccc@{}}
        \toprule
        \diagbox[width=28pt,height=22pt,innerleftsep=1pt,innerrightsep=1pt]{$S_j$}{$T_i$} & $(0.00,0.00)$ & $(3.43,0.00)$ & $(3.54,5.15)$ & $(0.00,5.15)$ \\
        \midrule
        \makecell{$\bm p_0$\\$\delta_*$} & \makecell{$(3.45,4.71)$\\$-2.57$} & \makecell{$(0.06,5.12)$\\$-1.29$} & \makecell{$(-0.34,0.34)$\\$-0.95$} & \makecell{$(3.29,-0.44)$\\$1.01$} \\
        \midrule
        \makecell{$\bm p_0$\\$\delta_*$} & \makecell{$(1.10,5.35)$\\$-0.81$} & \makecell{$(1.92,5.37)$\\$1.03$} & \makecell{$(2.55,-0.32)$\\$-0.69$} & \makecell{$(3.90,3.00)$\\$0.60$} \\
        \midrule
        \makecell{$\bm p_0$\\$\delta_*$} & \makecell{$(3.83,3.05)$\\$0.43$} & \makecell{$(-0.26,4.33)$\\$0.52$} & \makecell{$(-0.28,3.32)$\\$-0.77$} & \makecell{$(1.24,0.21)$\\$1.24$} \\
        \bottomrule
    \end{tabular*}
\end{table}
\begin{figure}[H]
    \centering
    \includegraphics[scale=0.8]{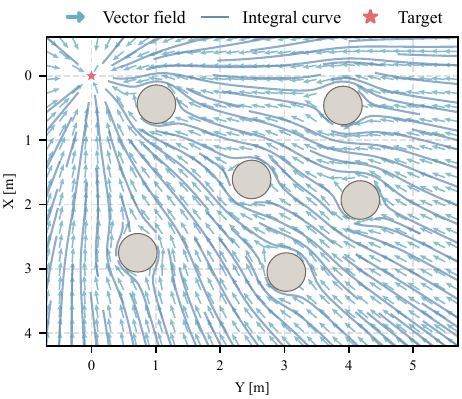}
    \caption{The nominal vector field with target $T_1$.}
    \label{fig:exp_vector_field}
\end{figure}
}

\afterpage{\unvbox\hwsetupbox}

\begin{figure*}[!b]
    \centering
    \newcommand{\expSnapshotPanel}[2]{\begin{tikzpicture}
            \node[anchor=south west, inner sep=0] (snapshot) at (0,0)
                {\includegraphics[width=\linewidth]{figs/generated/exp_snapshot_T#1.pdf}};
            \begin{scope}[overlay, x={(snapshot.south east)}, y={(snapshot.north west)},
                every node/.style={font=\scriptsize, inner sep=1pt, fill=none}]
                #2
            \end{scope}
        \end{tikzpicture}}
    \begin{minipage}{0.95\textwidth}
    \centering
    \includegraphics[width=\linewidth]{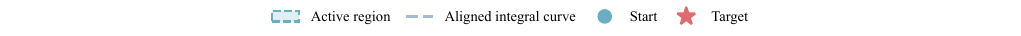}
    \par\smallskip
    \begin{subfigure}[t]{0.24\linewidth}
        \centering
        \expSnapshotPanel{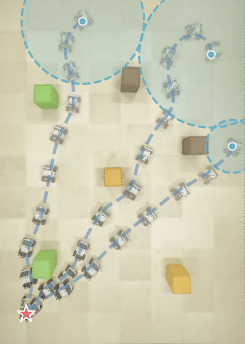}{\node[anchor=west, xshift=8pt] at (0.1033,0.0862) {$T_1$};
            \node[anchor=east, xshift=-5pt] at (0.89,0.82) {$S_1$};
            \node[anchor=west, xshift=5pt] at (0.330,0.9380) {$S_2$};
            \node[anchor=east, xshift=-5pt] at (0.98,0.5750) {$S_3$};
        }
        \caption{The first target.}
        \label{fig:target_1}
    \end{subfigure}
    \hfill
    \begin{subfigure}[t]{0.24\linewidth}
        \centering
        \expSnapshotPanel{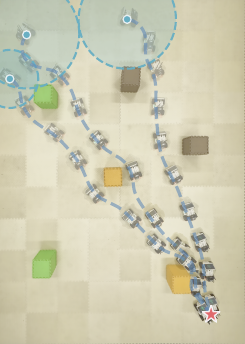}{\node[anchor=east, xshift=-8pt] at (0.8623,0.0865) {$T_2$};
            \node[anchor=west, xshift=5pt] at (0.10,0.8996) {$S_1$};
            \node[anchor=west, xshift=5pt] at (0.5192,0.9437) {$S_2$};
            \node[anchor=west, xshift=5pt] at (0.01,0.78) {$S_3$};
        }
        \caption{The second target.}
        \label{fig:target_2}
    \end{subfigure}
    \hfill
    \begin{subfigure}[t]{0.24\linewidth}
        \centering
        \expSnapshotPanel{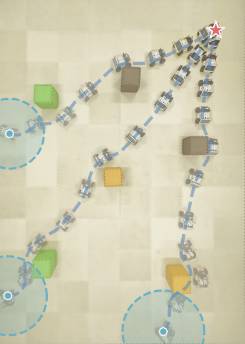}{\node[anchor=east, xshift=-8pt, yshift=4pt] at (0.8811,0.9115) {$T_3$};
            \node[anchor=west, xshift=5pt] at (0.0280,0.1393) {$S_1$};
            \node[anchor=south, yshift=5pt] at (0.64,0.0362) {$S_2$};
            \node[anchor=west, xshift=5pt] at (0.03,0.57) {$S_3$};
        }
        \caption{The third target.}
        \label{fig:target_3}
    \end{subfigure}
    \hfill
    \begin{subfigure}[t]{0.24\linewidth}
        \centering
        \expSnapshotPanel{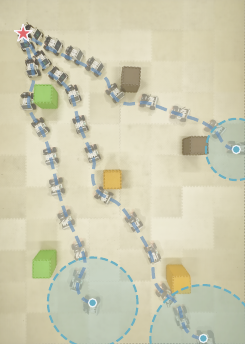}{\node[anchor=west, xshift=8pt, yshift=4pt] at (0.0936,0.9036) {$T_4$};
            \node[anchor=south, yshift=5pt] at (0.8306,0.0167) {$S_1$};
            \node[anchor=east, xshift=-5pt] at (0.99,0.54) {$S_2$};
            \node[anchor=west, xshift=5pt] at (0.3765,0.1195) {$S_3$};
        }
        \caption{The fourth target.}
        \label{fig:target_4}
    \end{subfigure}
    \end{minipage}

    \caption{Closed-loop experimental trajectories shown by compositing images of the UGV at successive time instants. Targets $T_1$--$T_4$ and initial configurations $S_1$--$S_3$ are listed in Table~\ref{tab:exp_setup}.}

    \label{fig:exp_snapshots}
\end{figure*}

{
Figure~\ref{fig:comparison_rms} further characterizes the variation in control effort across trials through interquartile ranges and 99th percentiles.
For ARVF at $k_{\bm \Omega}=0.02$, the low median control effort coexists with large 99th-percentile values, whereas NVF at $k_{\bm \Omega}=0.1$ exhibits large control effort across a broader portion of the distribution.
Both ARVF and NVF contain switching surfaces, whose crossings can produce large angular-velocity demands despite low feedback gains, as illustrated for ARVF in Figure~\ref{fig:case study continuity}.
At lower feedback gains, deviations from nominal integral curves can be larger, making the closed-loop trajectory more likely to cross switching surfaces and experience abrupt increases in angular velocity.
In contrast, DIVAL follows aligned integral curves that avoid the deformed switching surfaces, with control continuity guaranteed by Theorem~\ref{thm: smoothness}.
The absence of angular-velocity jumps helps explain the lower 99th-percentile values of $\alpha_{\max}$ observed for DIVAL relative to the lowest-gain baselines in the ARVF and NVF cases.
Within DIVAL, increasing the active-region radius and then optimizing the transition function successively narrows the interquartile ranges and lowers the 99th percentiles of all three control-effort metrics.
}

\begin{figure*}[!t]
    \centering
    \begin{minipage}[t]{\columnwidth}
        \captionsetup{type=figure}
        \centering
        \begin{subfigure}{0.9\columnwidth}
            \centering
            \includegraphics[width=\linewidth]{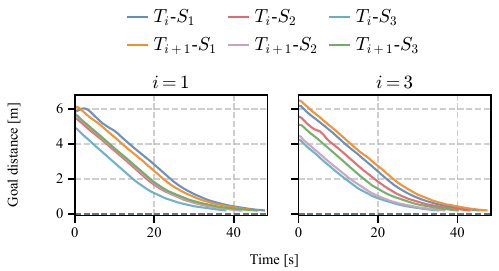}
            \caption{Distance to target.}
            \label{fig:exp_goal}
        \end{subfigure}
        \par\smallskip
        \begin{subfigure}{0.9\columnwidth}
            \centering
            \includegraphics[width=\linewidth]{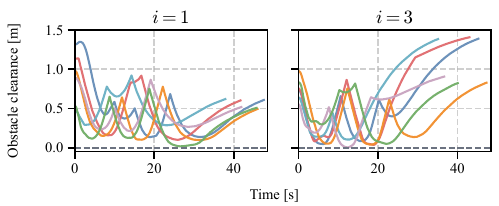}
            \caption{Clearance from the nearest obstacle.}
            \label{fig:exp_obstacle}
        \end{subfigure}
        \caption{Target distance and obstacle clearance of the $T_i$-$S_j$ trial, where $T_i$-$S_j$ denotes the trial from initial configuration $S_j$ to target $T_i$.
        Results for the first and last two targets appear on the left and right, respectively.
        }
        \label{fig:exp_metrics}
    \end{minipage}\hfill
    \begin{minipage}[t]{\columnwidth}
        \captionsetup{type=figure}
        \centering
        \begin{subfigure}{0.9\columnwidth}
            \centering
            \includegraphics[width=\linewidth]{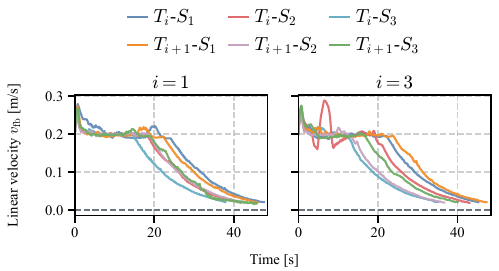}
            \caption{Linear velocity in \eqref{eq: commanded v}.}
        \end{subfigure}
        \par\smallskip
        \begin{subfigure}{0.9\columnwidth}
            \centering
            \includegraphics[width=\linewidth]{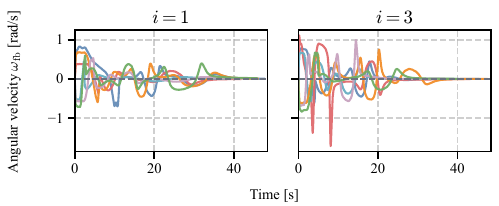}
            \caption{Angular velocity in \eqref{eq: commanded Omega}.}
        \end{subfigure}

        \caption{Control inputs of the UGV for the $T_i$-$S_j$ trial.}
        \label{fig:exp_control}
    \end{minipage}
\end{figure*}

\subsection{Hardware Experiments}
\label{subsec:hardware}

The hardware experiments assess how closely the UGV follows the aligned integral curves under real-world disturbances and whether it reaches the targets without collision.
We evaluate DIVAL on an AgileX LIMO UGV using an OptiTrack motion capture system for odometry measurements.
The workspace contains six static obstacles, each modeled as a disk with a radius of $\SI{0.3}{m}$ accounting for the vehicle footprint.
For each of four target positions at the workspace corners, we conduct three trials with distinct initial configurations, giving twelve trials in total, as summarized in Table~\ref{tab:exp_setup}.
Each trial terminates when the UGV is within $\SI{0.2}{m}$ of the target.

The nominal vector field used in the experiments is constructed using the method in \cite{Dalton1971potential} as illustrated in Figure~\ref{fig:exp_vector_field}, which is of class $C^\infty$ at non-singular positions \cite[Ch.~6.2]{batchelor2000introduction}.
For each trial, the active-region radius and transition function are selected according to \eqref{eq: max r_A} and \eqref{eq: lbd opt}.
By choosing $k = \min\!\left(v_{\max},k_d\|\bm p-\bm p_d\|\right)/\|\bm\eta\|$ as the scaling factor in \eqref{eq: v 2d}, the linear velocity becomes
\begin{equation*}
    \label{eq:scalar_gain_exp}
    v(\bm g)=\min\!\left(v_{\max},k_d\|\bm p-\bm p_d\|\right) \in C^0,
\end{equation*}
where $v_{\max}=\SI{0.2}{m/s}$, and $k_d=\SI{0.1}{s^{-1}}$.
{To compensate for disturbances in practice, the robust control law \eqref{eq: commanded inputs} uses gains $(k_v,k_h,k_p)=(1,3,3)$, and the resulting commands are sent to the UGV at $\SI{20}{Hz}$.}

{Under DIVAL with the robust control law, the experimental trajectories closely follow the aligned integral curves despite real-world disturbances, as shown in Figure~\ref{fig:exp_snapshots}.
Specifically, the distance from the UGV to the nearest point on the aligned integral curve has an RMS of $\SI{2.75}{cm}$ over all recorded position samples from the twelve trials.
The distance to target in Figure~\ref{fig:exp_metrics}(\subref{fig:exp_goal}) confirms that the UGV reaches the prescribed target neighborhood from different initial positions and heading mismatches.
The clearance of the UGV from the nearest obstacle remains positive as shown by Figure~\ref{fig:exp_metrics}(\subref{fig:exp_obstacle}).
Figure~\ref{fig:exp_control} presents the corresponding linear and angular velocity commands given by \eqref{eq: commanded inputs}.
These results demonstrate that the proposed method remains effective on the physical platform.}

\section{Conclusion}
\label{sec:conclusion}
In this paper, we have proposed the diffeomorphic vector-field alignment (DIVAL).
DIVAL aligns the nominal vector-field direction with the robot's initial heading by locally deforming the workspace, thereby enabling the robot to move along the deformed integral curve from the start.
We have shown that given a nominal vector field whose integral curves avoid obstacles and smoothly converge to the target, the closed-loop system under DIVAL also achieves collision-free convergence without inducing control discontinuities.
Compared with conventional vector-field-tracking baselines, DIVAL ensures safety and convergence in all trials at significantly lower control effort in Monte Carlo simulations.
Hardware experiments further examine the effectiveness of DIVAL against real-world disturbances.

{
Future directions include imposing curvature constraints on the reshaped vector field to accommodate robots with a prescribed trajectory curvature bound, and considering time-varying nominal vector fields for dynamic obstacle avoidance.
The difficulty in addressing the curvature constraint is that the curvature distribution of the reshaped vector field is determined by the nominal vector field and locally modified by DIVAL.
Consequently, a prescribed curvature bound over the entire workspace cannot generally be enforced through the deformation design alone.
This challenge calls for developing a nominal vector field that ensures collision-free convergence while satisfying a prescribed curvature bound for the obstacle environments considered in this paper.
Further research should also characterize how curvature changes under the DIVAL deformation to establish conditions under which the reshaped vector field preserves the properties of the nominal vector field, including the curvature bound.}

\appendices

\section{Calculation of Aligned Vector Field}
\label{appendix: eta derivation}
This appendix derives the inverse and Jacobian of the deformation map used to evaluate the aligned vector field in \eqref{eq: aligned vf}.
Knowing that the diffeomorphism \eqref{eq: diffeomorphism 3D} rotates each nominal position $\bm q$ by the rotation matrix $\exp([\bm{\Delta}(\bm{q})]_\times)$ around the initial position $\bm{p}_0$, we can write its inverse map as
\[
    \begin{aligned}
        &\Phi^{-1}(\bm{p})
        = \exp([\bm{\Delta}(\bm{q})]_\times)^{-1} (\bm{p} - \bm{p}_0) + \bm{p}_0,\\
        &\stackrel{\bm{\Delta}(\bm{q}) = \bm{\Delta}(\bm{p})}{=}\exp([\bm{\Delta}(\bm{p})]_\times)^\intercal (\bm{p} - \bm{p}_0) + \bm{p}_0,
    \end{aligned}
\]
where the second equality uses the radial symmetry of the transition function, namely, $\lambda(\bm{q}) = \lambda(\bm{p})$.

Next, the Jacobian of the diffeomorphism follows by differentiation:
\[
    \begin{aligned}
        \mathrm{d} \mathbf{\Phi}
        &= \exp([\bm \Delta]_\times) \mathrm{d} \bm{q} + \mathrm{d} (\exp([\bm \Delta]_\times)) \bar{\bm{q}} \\
        &= \exp([\bm \Delta]_\times) \mathrm{d} \bm{q} + \exp([\bm \Delta]_\times) \mathrm{d}( [\mathbf{J}_r \bm \Delta]_\times) \bar{\bm{q}}\\
        &= \exp([\bm \Delta]_\times) \mathrm{d} \bm{q} + \exp([\bm \Delta]_\times) \mathrm{d}([\bm \Delta]_\times) \bar{\bm{q}}\\
        &= \exp([\bm \Delta]_\times) \mathrm{d} \bm{q}
         + \exp([\bm \Delta]_\times) [\bm{\Delta}_* (\nabla \lambda^\intercal \mathrm{d}\bm{q})]_\times  \bar{\bm{q}} \\
        &= \exp([\bm \Delta]_\times) \mathrm{d} \bm{q}
         + \exp([\bm \Delta]_\times) [\bm{\Delta}_*]_\times  \bar{\bm{q}} (\nabla \lambda^\intercal \mathrm{d}\bm{q}) \\
        &= \underbrace{\exp([\bm \Delta]_\times)
         \left(
             \mathbf{I} + [\bm{\Delta}_*]_\times  \bar{\bm{q}} \nabla \lambda^\intercal
         \right)}_{\mathbf{J}_\Phi \ \text{in \eqref{eq: jac}}} \mathrm{d} \bm{q}
    \end{aligned}
\]
where $\mathbf{J}_r$ is the right Jacobian of $\exp([\bm \Delta]_\times) \in SO(3)$ and the third equality follows because $\bm{\Delta} \parallel \bm{\Delta}_*$ and hence $\mathbf{J}_r \bm \Delta = \bm \Delta $.
Substituting the above result into \eqref{eq: aligned vf} gives the aligned vector field.

\section{Calculation of Angular Velocity}
\label{appendix:control_inputs}
This appendix derives explicit expressions for the quantities needed to evaluate the angular-velocity law \eqref{eq: Omega}.
We first compute the Jacobian of the aligned vector field.
Let $\bm q=\Phi^{-1}(\bm p)$ and $\mathbf R_\Delta(\bm q)=\exp([\bm\Delta(\bm q)]_\times)$.
Differentiating the aligned vector field in \eqref{eq: aligned vf} gives the Jacobian $\mathbf{J}_{\bm \eta}$:
\begin{equation} \label{eq: J eta derivation}
    \begin{aligned}
        \mathbf J_{\bm{\eta}}(\bm p)
        &=
        \frac{\partial\bm{\eta}}{\partial\bm p}
        =
        \frac{\partial\bm{\eta}}{\partial\bm q}
        \frac{\partial\bm q}{\partial\bm p}\\
        &=
        \left(
        \frac{\partial}{\partial\bm q}
        \left[\mathbf J_\Phi(\bm q)\bm{\xi}(\bm q)\right]
        \right)
        \mathbf J_{\Phi^{-1}}(\bm p)\\
        &=
        \left(
        D\mathbf J_\Phi(\bm q)[\bm{\xi}]
        +\mathbf J_\Phi(\bm q)\mathbf J_{\bm{\xi}}(\bm q)
        \right)
        \mathbf J_{\Phi^{-1}}(\bm p)\\
        &=
        \mathbf R_\Delta(\Phi^{-1}(\bm p))
        \mathbf M_{\bm{\xi}}(\Phi^{-1}(\bm p))
        \mathbf J_{\Phi^{-1}}(\bm p),
    \end{aligned}
\end{equation}
where $D\mathbf J_\Phi(\bm q)[\bm{\xi}]$ is the directional derivative of $\mathbf J_\Phi$ along $\bm{\xi}$ and
the matrix $\mathbf M_{\bm{\xi}}$ is
\begin{equation*} \label{eq: M xi}
    \begin{aligned}
        \mathbf M_{\bm{\xi}}(\bm q)
        &=
        \mathbf J_{\bm{\xi}}
        +
        \left(
        \dot{\lambda}[\bm{\Delta}_*]_\times
        +[\bm{\Delta}_*]_\times\bm{\xi}\nabla\lambda^\intercal
        \right)\\
        &\quad
        +\dot{\lambda}[\bm{\Delta}_*]_\times^2
        \bar{\bm q}\nabla\lambda^\intercal\\
        &\quad
        +[\bm{\Delta}_*]_\times\bar{\bm q}
        \left(
        (\mathbf H_\lambda\bm{\xi})^\intercal
        +\nabla\lambda^\intercal\mathbf J_{\bm{\xi}}
        \right).
    \end{aligned}
\end{equation*}
In the above matrix, $\mathbf J_{\bm{\xi}}=\partial\bm{\xi}/\partial\bm q$ is the Jacobian of the nominal vector field, $\mathbf H_\lambda$ is the Hessian of the transition function, and $\dot{\lambda}(\bm q) \triangleq \nabla\lambda(\bm q)^\intercal\bm{\xi}(\bm q)$ is the directional derivative of the transition function along the nominal vector field.

Furthermore, since the term $\mathbf J_{\bm{\eta}}(\bm p)\bm{\eta}(\bm p)$ in \eqref{eq: Omega} is the acceleration of the evolution of the aligned integral curve, we denote it as $\bm a_{\bm{\eta}}(\bm p)$ and compute it as follows:
\begin{equation} \label{eq: acc eta}
    \begin{aligned}
        &\bm a_{\bm{\eta}}(\bm p)
        \triangleq
        \mathbf J_{\bm{\eta}}(\bm p)\bm{\eta}(\bm p)\\
        &=
        \mathbf R_\Delta(\bm q)\mathbf M_{\bm{\xi}}(\bm q)\bm{\xi}(\bm q)\\
        &=
        \mathbf R_\Delta
        \Big(
        \bm a_{\bm{\xi}}
        +2\dot{\lambda}[\bm{\Delta}_*]_\times\bm{\xi}
        +\dot{\lambda}^2[\bm{\Delta}_*]_\times^2
        \bar{\bm q}
        +\ddot{\lambda}[\bm{\Delta}_*]_\times\bar{\bm q}
        \Big),
    \end{aligned}
\end{equation}
where $\bm a_{\bm{\xi}}(\bm q)=\mathbf J_{\bm{\xi}}(\bm q)\bm{\xi}(\bm q)$ is the acceleration of the nominal integral curve's evolution, and
\[
    \ddot{\lambda}(\bm q)
            =
            \bm{\xi}(\bm q)^\intercal\mathbf H_\lambda(\bm q)\bm{\xi}(\bm q)
            +\nabla\lambda(\bm q)^\intercal\mathbf J_{\bm{\xi}}(\bm q)\bm{\xi}(\bm q).
\]
The four terms inside the parentheses of \eqref{eq: acc eta} are the nominal acceleration, the Coriolis term, the centripetal term, and the Euler term, respectively.

Finally, the term $\bm{\eta}(\bm p)\times\bm a_{\bm{\eta}}(\bm p)$ in \eqref{eq: Omega} is given by
\begin{equation} \label{eq: cross product eta}
    \begin{aligned}
        &\bm{\eta}(\bm p)\times\bm a_{\bm{\eta}}(\bm p)\\
        &=
        \mathbf R_\Delta
        \Big(
        \bm{\xi}\times\bm a_{\bm{\xi}}
        +C_\mathrm{rotation}\bm\Delta_*
        +C_\mathrm{radial}\bar{\bm q}\\
        &\quad\qquad +C_\mathrm{velocity}\bm{\xi}
        +C_\mathrm{cross}\bm\Delta_*\times\bar{\bm q}
        \Big),
    \end{aligned}
\end{equation}
where the scalar coefficients are
\begin{equation*} \label{eq: cross product coefficients}
    \begin{aligned}
        C_\mathrm{rotation}
        &=
        \dot{\lambda}
        \left(
        2\|\bm{\xi}\|^2-\left\langle\bm a_{\bm{\xi}},\bar{\bm q}\right\rangle
        \right)\\
        &\quad
        +3\dot{\lambda}^2
        \left\langle
        [\bm{\Delta}_*]_\times\bar{\bm q},\bm{\xi}
        \right\rangle
        +\dot{\lambda}^3
        \|[\bm{\Delta}_*]_\times\bar{\bm q}\|^2\\
        &\quad
        +\ddot{\lambda}\left\langle\bm{\xi},\bar{\bm q}\right\rangle,\\
        C_\mathrm{radial}
        &=
        \dot{\lambda}
        \left\langle\bm\Delta_*,\bm a_{\bm{\xi}}\right\rangle
        -\ddot{\lambda}
        \left\langle\bm\Delta_*,\bm{\xi}\right\rangle,\\
        C_\mathrm{velocity}
        &=
        -2\dot{\lambda}
        \left\langle\bm\Delta_*,\bm{\xi}\right\rangle,\\
        C_\mathrm{cross}
        &=
        -\dot{\lambda}^2
        \left\langle\bm\Delta_*,\bm{\xi}\right\rangle.
    \end{aligned}
\end{equation*}
Substituting \eqref{eq: cross product eta} into \eqref{eq: Omega} gives the angular velocity.

\section{Proof of Theorem~\ref{thm: trajectory relation}}
\label{appendix:proof_trajectory_relation}
The proof proceeds in three steps.
First, we establish that the control laws are well-defined along the aligned integral curve.
Then, we show that the proposed control laws maintain the alignment between the robot's heading and the vector field direction.
Finally, the resulting closed-loop trajectory is shown to be a time reparameterization of the aligned integral curve.

{
First, Theorem~\ref{thm:aligned vf} gives $\bm\eta\in C^r(\mathcal D_{\bm\eta})$, $r\ge1$, and $\bm\eta(\bm p_0)\neq\bm0$.
Let $\bm p_*$ be any singular point of $\bm\eta$.
The $C^r$ smoothness of $\bm\eta$ implies that its Jacobian $\mathbf J_{\bm\eta}$ is locally bounded, so there exists a Lipschitz constant $L>0$ such that
\[
    \|\bm\eta(\bm p)\|
    =\|\bm\eta(\bm p)-\bm\eta(\bm p_*)\|
    \le L\|\bm p-\bm p_*\|
\]
in a neighborhood of $\bm p_*$, where $\bm\eta(\bm p_*)=\bm0$.
Since the integral curve evolves according to $\frac{d}{d\tau}\psi_{\bm\eta}=\bm\eta(\psi_{\bm\eta})$, its distance to $\bm p_*$ cannot decrease faster than the bound proportional to its distance from $\bm p_*$, according to the above inequality.
Define $d(\tau)=\|\psi_{\bm\eta}(\tau;\bm p_0)-\bm p_*\|$ and consider a segment of the integral curve that remains in this neighborhood, starting at time $\tau_0$ with $d(\tau_0)>0$.
Where $d(\tau)>0$, the rate of change of this distance satisfies
\[
    \frac{d}{d\tau}d(\tau)
    \ge-\|\bm\eta(\psi_{\bm\eta}(\tau;\bm p_0))\|
    \ge-Ld(\tau).
\]
Integrating this inequality along the segment gives the positive lower bound
\[
    d(\tau)\ge d(\tau_0)e^{-L(\tau-\tau_0)}>0.
\]
This lower bound remains strictly positive at any finite endpoint of the segment, preventing the integral curve from reaching $\bm p_*$ in finite time.
Thus, $\bm\eta(\psi_{\bm\eta}(\tau;\bm p_0))\neq\bm0$ for all $\tau\in\mathcal T_{\bm\eta}$, and the control laws \eqref{eq: v} and \eqref{eq: Omega} are well-defined at configurations whose positions lie on this curve.
}

Second, define the set of configurations satisfying the alignment condition in \eqref{eq: control objective} as
\[
    \mathcal M
    \triangleq
    \left\{
    \bm g=(\bm p,\mathbf R)\in SE(3)\,\middle|\,
    \begin{aligned}
        &\bm p\in\mathcal D_{\bm\eta},\quad\bm\eta(\bm p)\neq\bm0,\\
        &\bm h=\hat{\bm\eta}(\bm p)
    \end{aligned}
    \right\}.
\]
For any $\bm g\in\mathcal M$, substituting the linear velocity \eqref{eq: v} into the translational kinematics in \eqref{eq:kinematics} gives
\[
    \dot{\bm p}
    =
    k(\bm g)\|\bm\eta(\bm p)\|\hat{\bm\eta}(\bm p)
    =
    k(\bm g)\bm\eta(\bm p).
\]
Moreover, for any $\bm g\in\mathcal M$, the time derivative of the aligned vector field direction in \eqref{eq: eta direction rate} becomes
\[
    \frac{d}{dt}\hat{\bm\eta}(\bm p(t))
    =
    k(\bm g)
    \big(
        \mathbf I
        -
        \hat{\bm\eta}(\bm p)\hat{\bm\eta}^{\intercal}(\bm p)
    \big)
    \mathbf J_{\bm\eta}(\bm p)\hat{\bm\eta}(\bm p).
\]
Next, directly substituting the angular velocity \eqref{eq: Omega} into the heading kinematics \eqref{eq: dot h} on the set $\mathcal M$ gives
\[
\begin{aligned}
    \dot{\bm h}
    &=
    {\bm \Omega}_{\mathrm{world}}\times\hat{\bm\eta}(\bm p)\\
    &=
    k(\bm g)
    \left(
        \frac{
            \bm\eta(\bm p)
            \times
            \mathbf J_{\bm\eta}(\bm p)\bm\eta(\bm p)
        }{
            \|\bm\eta(\bm p)\|^2
        }
    \right)
    \times
    \hat{\bm\eta}(\bm p)\\
    &=
    k(\bm g)
    \big(
        \hat{\bm\eta}(\bm p)
        \times
        \mathbf J_{\bm\eta}(\bm p)\hat{\bm\eta}(\bm p)
    \big)
    \times
    \hat{\bm\eta}(\bm p)\\
    &=
    k(\bm g)
    \big(
        \mathbf I
        -
        \hat{\bm\eta}(\bm p)\hat{\bm\eta}^{\intercal}(\bm p)
    \big)
    \mathbf J_{\bm\eta}(\bm p)\hat{\bm\eta}(\bm p).
\end{aligned}
\]
The last equality follows from the vector triple-product identity and $\|\hat{\bm\eta}(\bm p)\|=1$.
Therefore,
\[
    \dot{\bm h}
    =
    \frac{d}{dt}\hat{\bm\eta}(\bm p(t)),
    \quad \forall \bm g \in \mathcal M,
\]
which shows that $\mathcal M$ is invariant under the control laws \eqref{eq: v} and \eqref{eq: Omega}.
Since Theorem~\ref{thm:aligned vf} ensures that $\bm g_0\in\mathcal M$, the closed-loop system satisfies $\bm h(t)=\hat{\bm\eta}(\bm p(t))$ throughout its interval of existence.
Thus, \eqref{eq: control objective} is satisfied.

Finally, with the heading alignment established, the closed-loop system's heading $\bm h(t)$ coincides with the aligned vector field direction $\hat{\bm\eta}(\bm p(t))$ at every position.
Note that the speed scaling factor $k(\bm g)>0$ in the translational kinematics $\dot{\bm p} = k(\bm g)\bm\eta(\bm p)$ only changes how fast the robot moves along the aligned integral curve.
Define the time reparameterization
\[
    \tau(t)
    =
    \int_0^t k(\bm g(s;\bm g_0))\,\mathrm ds.
\]
Since $k(\bm g)>0$, $\tau(t)$ is strictly increasing.
Moreover, the chain rule gives
\[
\begin{aligned}
    \frac{d}{dt}\psi_{\bm\eta}(\tau(t);\bm p_0)
    &=
    \dot{\tau}(t)
    \bm\eta(\psi_{\bm\eta}(\tau(t);\bm p_0))\\
    &=
    k(\bm g(t;\bm g_0))
    \bm\eta(\psi_{\bm\eta}(\tau(t);\bm p_0)).
\end{aligned}
\]
Thus, $\psi_{\bm\eta}(\tau(t);\bm p_0)$ solves the same initial-value
problem as the closed-loop trajectory governed by $\dot{\bm p} = k(\bm g)\bm\eta(\bm p)$.
By uniqueness of solutions, the closed-loop trajectory satisfies
\[
    \bm p(t;\bm g_0)
    =
    \psi_{\bm\eta}(\tau(t);\bm p_0).
\]
This proves Theorem~\ref{thm: trajectory relation}.

\section{Proof of Theorem~\ref{thm: safety and convergence}}
\label{appendix:proof_safety_and_convergence}
The proof first shows that the control laws remain well-defined along the closed-loop trajectory for all $t\ge0$, then proves safety and convergence, respectively.

{
First, Assumption~\ref{ass:nominal safety and convergence} and the integral-curve relation \eqref{eq: image} established in Theorem~\ref{thm:aligned vf} ensure that the aligned integral curve $\psi_{\bm \eta}(\tau;\bm p_0)$ is defined for all $\tau\ge0$.
The upper bound on the scaling factor and \eqref{eq: time reparam} give \(\tau(t)\le k_{\max}t\), so \(\tau(t)\) remains finite for every finite \(t\).
By the time reparameterization in \eqref{eq: trajectory relation}, the closed-loop trajectory $\bm p(t;\bm g_0)$ is defined for all $t\ge0$.
The well-definedness argument in the proof of Theorem~\ref{thm: trajectory relation} ensures that the control laws \eqref{eq: v} and \eqref{eq: Omega} remain well-defined along the closed-loop trajectory for all \(t\in [0,\infty)\).
}

Second, the radius condition ensures that $\mathcal A\subseteq\mathcal F$.
The proof of Theorem~\ref{thm:aligned vf} shows that $\Phi(\mathcal A)=\mathcal A$ and that $\Phi$ is the identity outside $\mathcal A$, so $\Phi(\mathcal F)=\mathcal F$.
Combining the nominal safety property in Assumption~\ref{ass:nominal safety and convergence} with Theorem~\ref{thm: trajectory relation} and \eqref{eq: image} gives
\begin{equation*}
    \bm p(t;\bm g_0)
    =
    \Phi\left(\psi_{\bm\xi}(\tau(t);\bm p_0)\right)
    \in\mathcal F,\qquad \forall t\ge0,
\end{equation*}
which proves the safety of the closed-loop trajectory.

Finally, $\bm g_0\in\mathcal G_f$ implies $\bm p_0\in\mathcal R_{\bm\xi}^{a}$, and Assumption~\ref{ass:nominal safety and convergence} gives
\begin{equation*}
    \lim_{\tau\to\infty}
    \mathrm{dist}\left(\psi_{\bm\xi}(\tau;\bm p_0),\mathcal L_{\bm\xi}^{+}\right)=0.
\end{equation*}
The positive lower bound on the scaling factor yields $\tau(t)\ge k_{\min}t$, and hence $\tau(t)\to\infty$ as $t\to\infty$. Therefore,
\begin{equation}\label{eq: nominal convergence appendix}
    \lim_{t\to\infty}
    \mathrm{dist}\left(\psi_{\bm\xi}(\tau(t);\bm p_0),\mathcal L_{\bm\xi}^{+}\right)=0.
\end{equation}
Since the target limit set $\mathcal L_{\bm\xi}^{+}$ is either an equilibrium or a stable limit cycle, it is compact.
The continuity of $\Phi$ and the compactness of $\mathcal L_{\bm\xi}^{+}$ ensure that the distance in \eqref{eq: nominal convergence appendix} still tends to zero after both the nominal integral curve and the target limit set are deformed by $\Phi$.
Using $\bm p(t;\bm g_0)=\Phi(\psi_{\bm\xi}(\tau(t);\bm p_0))$ and $\mathcal L_{\bm \eta}^{+} = \Phi(\mathcal L_{\bm\xi}^{+})$, we can obtain
\begin{equation*}
    \lim_{t\to\infty}
    \mathrm{dist}\left(\bm p(t;\bm g_0),\mathcal L_{\bm \eta}^{+}\right)=0,
\end{equation*}
which proves the convergence of the closed-loop trajectory.

\section{Curvature of the Aligned Vector Field}
\label{appendix:curvature_bound}
{
This appendix derives the relation between the curvatures of the nominal and aligned vector fields.
Unless otherwise indicated, quantities related to the aligned vector field $\bm \eta$, including $\mathbf J_{\Phi^{-1}}$, $\bm a_{\bm \eta}$ and $\kappa_{\bm\eta}$, are evaluated at the deformed position $\bm p$, while other position-dependent quantities are evaluated at the nominal position $\bm q = \Phi^{-1}(\bm p)$.

First, with \eqref{eq: aligned vf} and \eqref{eq: cross product eta}, the curvature formula in \cite[Ch.~13.3]{Stewart2011Calculus} expands to
\begin{equation}
\label{eq: curvature bound split}
\begin{aligned}
    &\kappa_{\bm\eta}
    =\frac{\|\bm\eta\times\bm a_{\bm\eta}\|}{\|\bm\eta\|^3}
    =
    \frac{
        \left\|
        \bm\eta\times(\mathbf J_\Phi\bm a_{\bm\xi})
        +\bm\eta\times(\bm a_{\bm\eta}-\mathbf J_\Phi\bm a_{\bm\xi})
        \right\|
    }{\|\bm\eta\|^3}\\
    &\stackrel{\mathrm{(a)}}{=}
    \frac{
        1
    }{\|\bm\eta\|^3}
    \left\|
            \bm\eta\times
            \frac{\mathbf J_\Phi
            \left[(\bm\xi\times\bm a_{\bm\xi})\times\bm\xi\right]}
            {\|\bm\xi\|^2}
            +
            \bm\eta\times
            (\bm a_{\bm\eta}-\mathbf J_\Phi\bm a_{\bm\xi})
        \right\|
    \\
    &\le
    \frac{
        \left\|\mathbf J_\Phi
        \left[(\bm\xi\times\bm a_{\bm\xi})\times\bm\xi\right]\right\|
    }{
        \|\mathbf J_\Phi\bm\xi\|^2\|\bm\xi\|^2
    }+
    \frac{
        \|\bm a_{\bm\eta}-\mathbf J_\Phi\bm a_{\bm\xi}\|
    }{
        \|\mathbf J_\Phi\bm\xi\|^2
    },
\end{aligned}
\end{equation}
where step (a) follows by substituting the equation derived from the vector triple-product identity:
\[
\mathbf J_\Phi\bm a_{\bm\xi}
=\frac{\langle\bm\xi,\bm a_{\bm\xi}\rangle}{\|\bm\xi\|^2}\bm\eta
+\frac{\mathbf J_\Phi\left[(\bm\xi\times\bm a_{\bm\xi})\times\bm\xi\right]}{\|\bm\xi\|^2}.
\]

Next, we  derive bounds for the first and second terms on the right-hand side of \eqref{eq: curvature bound split}, respectively.
For the first term, with $\kappa_{\bm\xi}(\bm q)=\|\bm\xi\times\bm a_{\bm\xi}\|/\|\bm\xi\|^3$ denoting the nominal curvature, we have
\begin{equation}
\label{eq: curvature nominal term}
\begin{aligned}
    &\frac{
        \left\|\mathbf J_\Phi
        \left[(\bm\xi\times\bm a_{\bm\xi})\times\bm\xi\right]\right\|
    }{
        \|\mathbf J_\Phi\bm\xi\|^2\|\bm\xi\|^2
    }\le
    \frac{
        \|\mathbf J_\Phi\|\,
        \|(\bm\xi\times\bm a_{\bm\xi})\times\bm\xi\|
    }{
        \|\mathbf J_\Phi\bm\xi\|^2\|\bm\xi\|^2
    }\\
    &\stackrel{\bm\xi\perp(\bm\xi\times\bm a_{\bm\xi})}{=}
    \frac{
        \|\mathbf J_\Phi\|\,
        \|\bm\xi\times\bm a_{\bm\xi}\|
    }{
        \|\mathbf J_\Phi\bm\xi\|^2\|\bm\xi\|
    }=
    \frac{
        \|\mathbf J_\Phi\|\,\|\bm\xi\|^2
    }{
        \|\mathbf J_\Phi\bm\xi\|^2
    }
    \kappa_{\bm\xi}(\bm q)\\
    &\stackrel{\mathrm{(b)}}{\le}
    \|\mathbf J_\Phi\|\,\|\mathbf J_{\Phi^{-1}}\|^2
    \kappa_{\bm\xi}(\bm q)
    \stackrel{\mathrm{(c)}}{\le}
    \underbrace{
        \left(
            1+\|\bm\Delta_*\|\,\|\bar{\bm q}\|\,\|\nabla\lambda(\bm q)\|
        \right)^3
    }_{\triangleq C_1(\bm q)}
    \kappa_{\bm\xi}(\bm q),
\end{aligned}
\end{equation}
where (b) uses $\|\bm\xi\|=\|\mathbf J_{\Phi^{-1}}\mathbf J_\Phi\bm\xi\|\le\|\mathbf J_{\Phi^{-1}}\|\,\|\mathbf J_\Phi\bm\xi\|$.
Step (c) uses the bounds on \eqref{eq: inv} and \eqref{eq: jac}, which are $ \|\mathbf J_\Phi\|$, $\|\mathbf J_{\Phi^{-1}}\| \le 1+\|\bm\Delta_*\|\,\|\bar{\bm q}\|\,\|\nabla\lambda(\bm q)\|$.

For the second term, substituting \eqref{eq: acc eta} yields
\begin{equation}
\label{eq: curvature bound C2}
\begin{aligned}
    &\frac{
        \|\bm a_{\bm\eta}-\mathbf J_\Phi\bm a_{\bm\xi}\|
    }{
        \|\mathbf J_\Phi\bm\xi\|^2
    }\\
    &=
    \frac{
        \left\|
            2\dot\lambda[\bm\Delta_*]_\times\bm\xi
            +
            (\bm\xi^\intercal\mathbf H_\lambda\bm\xi)
            [\bm\Delta_*]_\times\bar{\bm q}
            +
            \dot\lambda^2[\bm\Delta_*]_\times^2\bar{\bm q}
        \right\|
    }{
        \|\mathbf J_\Phi\bm\xi\|^2
    }\\
    &\stackrel{\mathrm{(d)}}{\le}
    \frac{\|\bm\xi\|^2}{\|\mathbf J_\Phi\bm\xi\|^2}
    \|\bm\Delta_*\|
    \Big[
        2\|\nabla\lambda\|
        +\|\bar{\bm q}\|\,\|\mathbf H_\lambda\|
        +\|\bm\Delta_*\|\,\|\bar{\bm q}\|\,\|\nabla\lambda\|^2
    \Big]\\
    &\stackrel{\mathrm{(e)}}{\le}
    C_1(\bm q)^{\frac{2}{3}}\|\bm\Delta_*\|
    \Big[
        2\|\nabla\lambda\|
        +\|\bar{\bm q}\|\,\|\mathbf H_\lambda\|
        +\|\bm\Delta_*\|\,\|\bar{\bm q}\|\,\|\nabla\lambda\|^2
    \Big]\\
    &=
    C_1(\bm q)^{\frac{2}{3}}\|\bm\Delta_*\|\!
    \Big[
        2\|\nabla\lambda\|
        +\|\bar{\bm q}\|\,\|\mathbf H_\lambda\|
        +
        \left(C_1(\bm q)^{\frac{1}{3}}-1\right)
        \|\nabla\lambda\|
    \Big]\\
    &\stackrel{\mathrm{(f)}}{\le}
    C_1(\bm q)^{\frac{2}{3}}\|\bm\Delta_*\|
    \Big[
        2\|\nabla\lambda\|
        +\|\bar{\bm q}\|\,\|\mathbf H_\lambda\|\\
    &\qquad+
        \left(C_1(\bm q)^{\frac{1}{3}}-1\right)
        \left(
            2\|\nabla\lambda\|
            +\|\bar{\bm q}\|\,\|\mathbf H_\lambda\|
        \right)
    \Big]\\
    &=
    \underbrace{
        C_1(\bm q)\|\bm\Delta_*\|
        \left(
            2\|\nabla\lambda\|
            +\|\bar{\bm q}\|\,\|\mathbf H_\lambda\|
        \right)
    }_{\triangleq C_2(\bm q)}.
\end{aligned}
\end{equation}
The first equality uses the orthogonality of $\mathbf R_\Delta$.
Step (d) uses $|\dot\lambda|\le\|\nabla\lambda\|\,\|\bm\xi\|$ and $|\bm\xi^\intercal\mathbf H_\lambda\bm\xi|\le\|\mathbf H_\lambda\|\,\|\bm\xi\|^2$.
Step (e) reuses the scaling in step (b) and hence $\|\bm\xi\|^2/\|\mathbf J_\Phi\bm\xi\|^2 \le \| \mathbf J_{\Phi^{-1}} \|^2 \le C_1(\bm q)^{2/3}$.
Step (f) uses $C_1(\bm q)\ge1$ and $\|\nabla\lambda\|\le2\|\nabla\lambda\|+\|\bar{\bm q}\|\,\|\mathbf H_\lambda\|$.

Combining the bounds for the two terms in \eqref{eq: curvature bound split} establishes
\begin{equation}
\label{eq: curvature bound result}
    \kappa_{\bm\eta}(\Phi(\bm q))
    \le C_1(\bm q)\kappa_{\bm\xi}(\bm q)+C_2(\bm q),
\end{equation}
which proves the bound in \eqref{eq: aligned curvature}.
Since $C_1$ and $C_2$ are finite along the nominal integral curve, \eqref{eq: nominal curvature boundedness} in Assumption~\ref{ass:nominal safety and convergence} implies that the right-hand side of \eqref{eq: curvature bound result} is also finite.
Therefore, for any initial configuration $\bm g_0 = (\bm p_0, \mathbf{R}_0) \in \mathcal{G}_f$, the curvature along the aligned integral curve remains finite, i.e., $\sup_{t \in [0,\infty)} \kappa_{\bm \eta}(\psi_{\bm \eta}(t;\bm p_0)) < \infty$.

The bound in \eqref{eq: curvature bound result} is attained in two special cases.
At $\bm p=\bm p_0$, we have $\bm q=\bm p_0$, $\bar{\bm q}=\bm 0$, and $\nabla\lambda=\bm 0$.
For $\bm p\notin\mathcal A$, we have $\bm q=\bm p$ and $\lambda=0$, $\nabla\lambda=\bm 0$, and $\mathbf H_\lambda=\mathbf 0$.
In both cases, $C_1(\bm q)=1$ and $C_2(\bm q)=0$, $\mathbf J_\Phi$ is a rotation matrix and $\bm a_{\bm\eta}=\mathbf J_\Phi\bm a_{\bm\xi}$, so the inequalities in \eqref{eq: curvature bound split} hold with equality.
Consequently, $\kappa_{\bm\eta}(\bm p)=\kappa_{\bm\xi}(\bm p)$ at the initial position and outside the active region.

Finally, consider varying $r_{\mathcal A}$ with a fixed initial heading mismatch, a fixed transition function, and a fixed relative position $\bar{\bm q}/r_{\mathcal A}$.
Under this spatial rescaling, the chain rule gives
\[
\|\nabla\lambda\| \ \propto \ r_{\mathcal A}^{-1},\qquad
\|\mathbf H_\lambda\| \ \propto  \ r_{\mathcal A}^{-2},\qquad
\|\bar{\bm q}\| \ \propto \ r_{\mathcal A}.
\]
Therefore, $C_1$ in \eqref{eq: curvature nominal term} with the product $\|\bar{\bm q}\|\,\|\nabla\lambda\|$ is unchanged.
In the expression for $C_2$ in \eqref{eq: curvature bound C2}, both $\|\nabla\lambda\|$ and $\|\bar{\bm q}\|\,\|\mathbf H_\lambda\|$ scale as $r_{\mathcal A}^{-1}$.
The prefactor $C_1\|\bm\Delta_*\|$ remains unchanged, and hence $C_2$ scales as $r_{\mathcal A}^{-1}$ at corresponding relative positions.
}

\bibliographystyle{IEEEtran}
\bibliography{mybibfile}
\end{document}